\documentclass[smallextended]{svjour3}       % onecolumn (second format)
\smartqed  % flush right qed marks, e.g. at end of proof
\RequirePackage{fix-cm}
\usepackage{multirow}
\usepackage{pifont}   %for the \cmark and \xmark
\newcommand{\cmark}{\ding{51}}%
\newcommand{\xmark}{\ding{55}}%
\usepackage[utf8]{inputenc}
\usepackage{comment}
\usepackage{appendix}
\usepackage{graphicx}
\usepackage{xcolor}
\usepackage{amssymb}
\usepackage{amsmath}
\usepackage{mathtools}
\usepackage{commath}
\usepackage{url}
\usepackage{adjustbox}
\usepackage{booktabs}
\usepackage{ulem}
\usepackage[justification=centering]{caption}
\newtheorem{mydef}{Definition}
\usepackage{subfigure}

\usepackage{xspace}

\newcommand{\gm}[1]{\textcolor{black}{#1}\xspace}

\begin{document}

\title{Deep Graph Similarity Learning: A Survey %\thanks{Grants or other notes
}


\author{Guixiang Ma         \and
        Nesreen K. Ahmed    \and    Theodore L. Willke  \and    Philip S. Yu
        %\footnote{Philip S. Yu is supported by NSF under grants III-1526499, III-1763325, III-1909323, and SaTC-1930941.}  %etc.
}

\authorrunning{Ma, Ahmed, Willke, and Yu} % if too long for running head

\institute{G. Ma \at
              Intel Labs, Intel Corporation, Hillsboro, OR 97124, USA \\
              \email{guixiang.ma@intel.com}           %  \\
           \and
           N. K. Ahmed \at
              Intel Labs, Intel Corporation, Santa Clara, CA 95054, USA \\
              \email{nesreen.k.ahmed@intel.com}
            \and
            T. Willke \at
               Intel Labs, Intel Corporation, Hillsboro, OR 97124, USA \\
              \email{ted.willke@intel.com}   
            \and
            P. Yu \at
                Department of Computer Science, University of Illinois at Chicago, IL 60607, USA \\
                \email{psyu@uic.edu}
}

\date{}

\maketitle

\begin{abstract}
In many domains where data are represented as graphs, learning a similarity metric among graphs is considered a key problem, which can further facilitate various learning tasks, such as classification, clustering, and similarity search. 
Recently, there has been an increasing interest in deep graph similarity learning, where the key idea is to learn a deep learning model that maps input graphs to a target space such that the distance in the target space approximates the structural distance in the input space. 
% In recent years, with the development of deep learning techniques, there has emerged a fair amount of deep graph similarity learning strategies, for example the graph neural network (GNN) based graph similarity learning methods, for various applications. 
Here, we provide a comprehensive review of the existing literature of deep graph similarity learning. We propose a systematic taxonomy for the methods and applications.  Finally, we discuss the challenges and future directions for this problem. 

\keywords{Metric learning, Similarity learning, Graph Neural Networks, Graph Convolutional Networks, Higher-order Networks, Graph Similarity, Structural Similarity, Graph Matching, Deep graph similarity learning}
% \keywords{Graph Similarity \and Metric Learning \and Deep Learning \and Deep Graph Models}
\end{abstract}

\section{Introduction}
\label{sec:intro}
Learning an adequate similarity measure on a feature space can significantly determine the performance of machine learning methods. Learning such measures automatically from data is the primary aim of similarity learning. 
Similarity/Metric learning refers to learning a function to measure the distance or similarity between objects, which is a critical step in many machine learning problems, such as classification, clustering, ranking, etc. For example, in k-Nearest Neighbor (kNN) classification \cite{cover1967nearest}, a metric is needed for measuring the distance between data points and identifying the nearest neighbors; in many clustering algorithms, similarity measurements between data points are used to determine the clusters. Although there are some general metrics like Euclidean distance that can be used for getting similarity measure between objects represented as vectors, these metrics often fail to capture the specific characteristics of the data being studied, especially for structured data. Therefore, it is essential to find or learn a metric for measuring the similarity of data points involved in the specific task. 

Metric learning has been widely studied in many fields on various data types. For instance, in computer vision, metric learning has been explored on images or videos for image classification, object recognition, visual tracking, and other learning tasks \cite{mensink2012metric,guillaumin2009you,jiang2012order}. In information retrieval, such as in search engines, metric learning has been used to determine the ranking of relevant documents to a given query \cite{lee2008rank,lim2013robust}. In this paper, we survey the existing work in similarity learning for graphs, which encode relational structures and are ubiquitous in various domains. 

Similarity learning for graphs has been studied for many real applications, such as molecular graph classification in chemoinformatics \cite{horvath2004cyclic,frohlich2006kernel}, protein-protein interaction network analysis for disease prediction \cite{borgwardt2007graph}, binary function similarity search in computer security \cite{li2019graph}, multi-subject brain network similarity learning for neurological disorder analysis \cite{ktena2018metric}, etc. In many of these application scenarios, the number of training samples available is often very limited, making it a difficult problem to directly train a classification or prediction model. With graph similarity learning strategies, these applications benefit from pairwise learning that utilizes every pair of training samples to learn a metric for mapping the input data to the target space, which further facilitates the specific learning task. 

In the past few decades, many techniques have emerged for studying the similarity of graphs. Early on, multiple graph similarity metrics were defined, such as the Graph Edit Distance \cite{bunke1983inexact}, Maximum Common Subgraph \cite{bunke1998graph,wallis2001graph}, and Graph Isomorphism \cite{dijkman2009graph,berretti2001efficient}, to address the problem of graph similarity search and graph matching. However, the computation of these metrics is an NP-complete problem in general \cite{zeng2009comparing}. Although some pruning strategies and heuristic methods have been proposed to approximate the values and speed up the computation, it is difficult to analyze the computational complexities of the above heuristic algorithms and the sub-optimal solutions provided by them are also unbounded \cite{zeng2009comparing}. Therefore, these approaches are feasible only for graphs of relatively small size and in practical applications where these metrics are of primary interest. Thus it is hard to adapt these methods to new tasks. In addition, for other methods that are relatively more efficient like the Weisfeiler-Lehman method in \cite{douglas2011weisfeiler}, since it is developed specifically for isomorphism testing without mapping functions, it cannot be applied for general graph similarity learning. More recently, researchers have formulated similarity estimation as a learning problem where the goal is to learn a model that maps a pair of graphs to a similarity score based on the graph representations. For example, graph kernels, such as path-based kernels \cite{borgwardt2005shortest} and the subgraph matching kernel \cite{yan2005substructure,yoshida2019learning}, were proposed for graph similarity learning.  Traditional graph embedding techniques, such as geometric embedding, are also leveraged for graph similarity learning \cite{johansson2015learning}. 

\gm{With the emergence of deep learning techniques, graph neural networks (GNNs) have become a powerful new tool for learning representations on graphs with various structures for various tasks. The main distinction between GNNs and the traditional graph embedding is that GNNs address graph-related tasks in an end-to-end manner, where the representation learning and the target learning task are conducted jointly \cite{wu2020comprehensive}, while the graph embedding generally learns graph representations in an isolated stage and the learned representations are then used for the target task. Therefore, the GNN deep models can better leverage the graph features for the specific learning task compared to the graph embedding methods. Moreover, GNNs are easily adapted and extended for various graph related tasks, including deep graph similarity learning tasks in different domains. For instance, in brain connectivity network analysis in neuroscience, community structure among the nodes (i.e. brain regions) within the brain network is an important factor that should be considered when learning node representations for cross-subject similarity analysis. However, none of the traditional graph embedding methods are able to capture such special structure and jointly leverage the learned node representations for similarity learning on brain networks. In \cite{ma2019similarity}, a higher-order GNN model is developed to encode the community-structure of brain networks during the representation learning and leverage it for the similarity learning task on these brain networks. Some more examples from other domains include the GNN-based graph similarity predictive models introduced for chemical compound queries in computational chemistry \cite{bai2019simgnn}, and the deep graph matching networks proposed for binary function similarity search and malware detection in computer security \cite{li2019graph,ijcai2019-522}.}

In this survey paper, we provide a systematic review of the existing work in deep graph similarity learning. Based on the different graph representation learning strategies and how they are leveraged for the deep graph similarity learning task, we propose to categorize deep graph similarity learning models into three groups: Graph Embedding based-methods, GNN-based methods, and Deep Graph Kernel-based methods. Additionally, we sub-categorize the models based on their properties. Table ~\ref{tab:taxonomy} shows our proposed taxonomy, with some example models for each category as well as the relevant applications. In this survey, we will illustrate how these different categories of models approach the graph similarity learning problem. We will also discuss the loss functions used for the graph similarity learning task. 

\paragraph{Scope and Contributions.} This paper is focused on surveying the recently emerged deep models for graph similarity learning, where the goal is to use deep strategies on graphs for learning the similarity of given pairs of graphs, instead of computing similarity scores based on predefined measures. We emphasize that this paper does not attempt to survey the extensive literature on graph representation learning, graph neural networks, and graph embedding. Prior work has focused on these topics (see~\cite{cai2018comprehensive,goyal2018graph,lee2019attention,wu2019comprehensive,rossi2019community,cui2018survey,zhang2018network} for examples). Here instead, we focus on deep graph representation learning methods that explicitly focus on modeling graph similarity. To the best of our knowledge, this is the first survey paper on this problem. We summarize the main contributions of this paper as follows:
\begin{itemize}
    \item[$\circ$] Two comprehensive taxonomies to categorize the literature of the emerging field of deep graph similarity learning, based on the type of models and the type of features adopted by the existing methods, respectively. 
    \item[$\circ$] Summary and discussion of the key techniques and building blocks of the models in each category.
    \item[$\circ$] Summary and comparison of the different deep graph similarity learning models across the taxonomy.
    \item[$\circ$] Summary and discussion of the real-world applications that can benefit from deep graph similarity learning in a variety of domains. 
    \item[$\circ$] Summary and discussion of the major challenges for deep graph similarity learning, the future directions, and the open problems.  
    
\end{itemize}
% \textcolor{red}{\textbf{TODO:} Add contribution list.}

\paragraph{Organization.} The rest of the paper is organized as follows. In Section~\ref{sec:prelims}, we introduce notation, preliminary concepts, and define the graph similarity learning problem. In Section 3, we introduce the taxonomy with detailed illustrations of the existing deep models. In Section 4, we summarize the datasets and evaluations adopted in the existing works. In Section~\ref{sec:apps}, we present the applications of deep graph similarity learning in various domains. In Section~\ref{sec:challenges}, we discuss the remaining challenges in this area and highlight future directions. Finally, we conclude in Section~\ref{sec:conc}. 

\section{Notation and Preliminaries}\label{sec:prelims}

In this section, we provide the necessary notation and definitions of the fundamental concepts pertaining to the graph similarity problem that will be used throughout this survey.  The notation is summarized in Table~\ref{tab:notation}.

\begin{table*}
% \small
\begin{center}
\caption{Summary of Notation}
\label{tab:notation}
\begin{tabular}{lllll}
\toprule
% Notations & Descriptions\\
% \midrule
$G$ & & & & Input graph\\
$V$& & & & The set of nodes in a graph $G$\\
$E$& & & & The set of edges in a graph $G$\\
$a,\mathbf{a},\mathbf{A}$& & & & Scalar, vector, matrix\\

$\mathcal{G}$ & & & & Graph set $\mathcal{G} = \{G_1,G_2,\cdots ,G_n\}$\\
$\mathcal{M}$ & & & & Similarity function\\
$s_{ij}$ & & & & Similarity score between two graphs $G_i , G_j \in \mathcal{G}$\\

$\mathbb{R}^{m\times m}$& & & & $m-$dimensional Euclidean space\\
$\mathbf{I}_m$& & & & Identity matrix of dimension $m$ \\
$\mathbf{A}^T$& & & & Matrix transpose\\
$\mathbf{L}$& & & & Laplacian matrix\\
$g_\theta*\mathbf{x}$& & & & Convolution of $g_\theta$ and $\mathbf{x}$ \\

% $\mathcal{X}$ & each calligraphic letter represents a tensor \\
% $\left\langle \cdot, \cdot \right\rangle$ & denotes inner product\\
% $\circ$ & denotes tensor product (outer product)\\
% $\otimes$ & denotes Kronecker product\\
% $\odot$ & denotes Khatri-Rao product\\
\bottomrule
\end{tabular}
\end{center}
\vspace{-3mm}
\end{table*}

Let $G = (V,E,\mathbf{A})$ denote a graph, where $V$ is the set of nodes, $E \subseteq V \times V$ is the set of edges, and $\mathbf{A} \in \mathbb{R}^{|V| \times |V|}$ is the adjacency matrix of the graph. This is a general notation for graphs that covers different types of graphs, including unweighted/weighted graphs, undirected/directed graphs, and attributed/non-attributed graphs.

We are also assuming a set of graphs as input, $\mathcal{G} = \{G_1, G_2, \dots, G_n\}$, and the goal is measure/model their pairwise similarity. This relates to the classical problem of graph isomorphism and its variants. In graph isomorphism~\cite{miller1979graph}, two graphs $G = (V_G,E_G)$ and $H = (V_H,E_H)$ are isomorphic (i.e., $G \cong H$), if there is a \gm{mapping} function $\pi: V_G \rightarrow V_H$, such that $(u,v) \in E_G$ iff $(\pi(u),\pi(v)) \in E_H$. The graph isomorphism is an NP problem, and no efficient algorithms are known for it. Subgraph isomorphism is a generalization of the graph isomorphism problem. In subgraph isomorphism, the goal is to answer for two input graphs $G$ and $H$, if there is a subgraph of $G$ ($G' \subset G$) such that $G'$ is isomorphic to $H$ (i.e., $G' \cong H$). This is suitable in a setting in which the two graphs have different sizes. The subgraph isomorphism problem has been proven to be NP-complete (unlike the graph isomorphism problem)~\cite{Garey1978ComputersAI}. The maximum common subgraph problem is another less-restrictive measure of graph similarity, in which the similarity between two graphs is defined based on the size of the largest common subgraph in the two input graphs. However, this problem is also NP-complete~\cite{Garey1978ComputersAI}.

\begin{mydef}[\textbf{Graph Similarity Learning}]\label{def:gsim}
Let $\mathcal{G}$ be an input set of graphs, $\mathcal{G} = \{G_1,G_2,\cdots ,G_n\}$ where $G_i=(V_i, E_i, \mathbf{A}_i)$. Let $\mathcal{M}$ denote a learnable similarity function, such that $\mathcal{M}: (G_i, G_j) \rightarrow \mathbb{R}$, for any pair of graphs $G_i, G_j \in \mathcal{G}$. Assume $s_{ij} \in \mathbb{R}$ denote the similarity score computed using $\mathcal{M}$ between pairs $G_i$ and $G_j$. Then $\mathcal{M}$ is symmetric if and only if $s_{ij} = s_{ji}$ for any pair of graphs $G_i, G_j \in \mathcal{G}$. $\mathcal{M}$ should satisfy the property that: $s_{ii} >= s_{ij}$ for any pair of graphs $G_i, G_j \in \mathcal{G}$. And, $s_{ij}$ is minimum if $G_i$ is the complement of $G_j$, i.e, $G_i = \Bar{G_j}$, for any graph $G_j \in \mathcal{G}$.
\end{mydef}

Clearly, graph isomorphism and its related variants (e.g., subgraph isomorphism, maximum common subgraphs, etc.) are focused on measuring the topological equivalence of graphs, which gives rise to a binary similarity measure that outputs $1$ if two graphs are isomorphic and $0$ otherwise. While these methods may sound intuitive, they are actually more restrictive and difficult to compute for large graphs. Here instead, we focus on a relaxed notion of graph similarity that can be measured using machine learning models, where the goal is to learn a model that quantifies the degree of structural similarity and relatedness between two graphs. This is slightly similar to the work done on modeling the structural similarity between nodes in the same graph~\cite{ahmedrole2020,rossi2014role,ahmed2018learning}. We formally state the definition of graph similarity learning (GSL) in Definition~\ref{def:gsim}. Note that in the case of deep graph similarity learning, the similarity function $\mathcal{M}$ is a neural network model that can be trained in an end-to-end fashion. 

\section{Taxonomy of Models}
\begin{figure}[t]
\centering
    \begin{subfigure}[ ]{		\includegraphics[width=.47\linewidth] {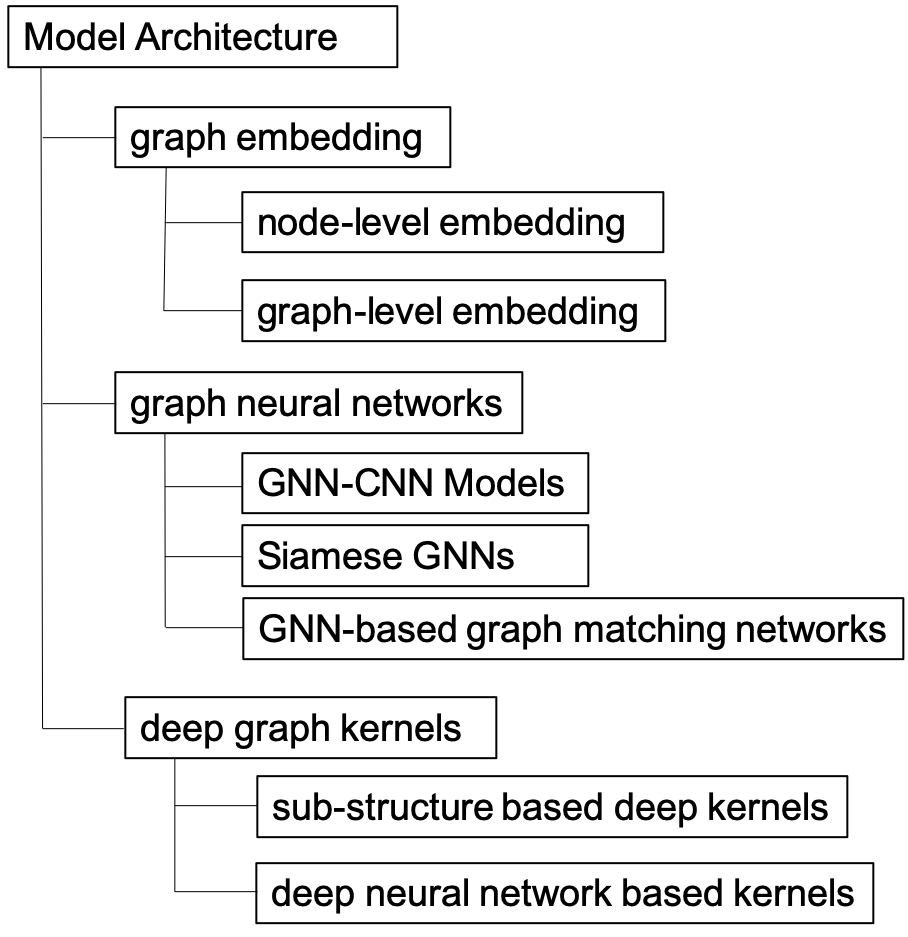}
		    \label{fig:model_taxonomy}
		}%
		\end{subfigure}
		\begin{subfigure}[ ]{
		\includegraphics[width=.43\linewidth]{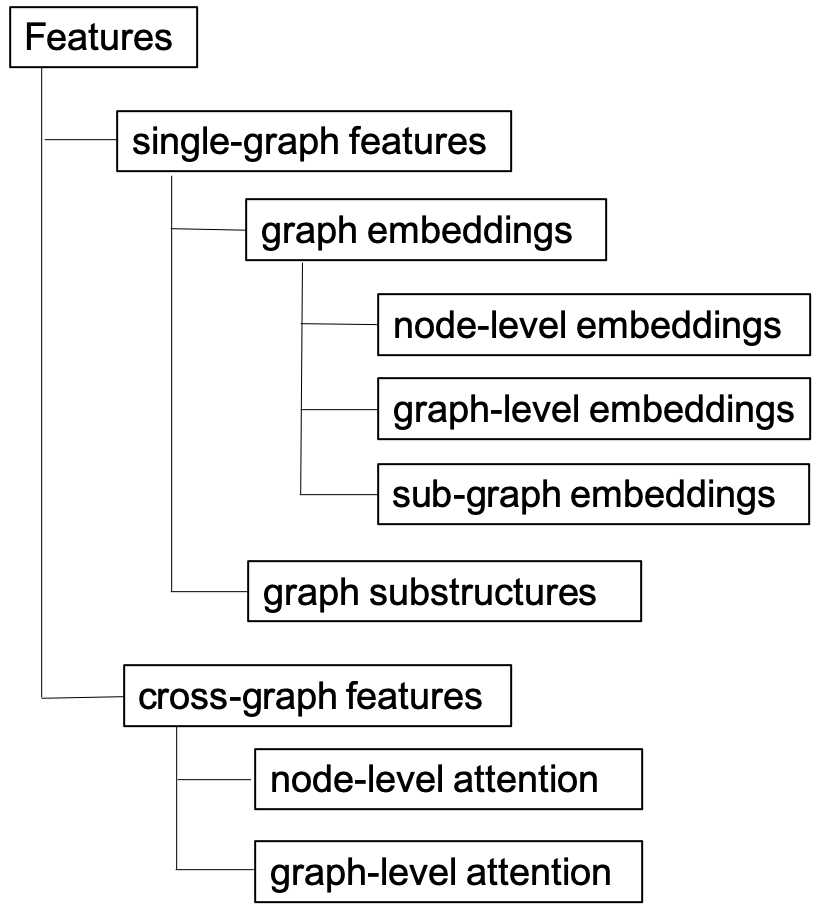}
		\label{fig:feature_taxonomy}
		}%
		\end{subfigure}  
  \caption{Proposed taxonomy for categorizing the literature of deep graph similarity learning based on (a) Model architecture, (b) Type of features.}
  \label{fig:taxonomy}
\end{figure}

% \usepackage{multirow}
\setlength{\tabcolsep}{5.2pt} % General space between cols (6pt standard)
\renewcommand{\arraystretch}{1.3}

{

\begin{table*}[]
\begin{center}
\caption{A Taxonomy of Deep Graph Similarity Learning Methods}
\label{tab:taxonomy}
% \begin{adjustbox}{angle=90}
\scalebox{0.8}{
\small
\begin{tabular}{c|l|l|c|c|c|c|c|l}
\toprule
\multicolumn{2}{c|}{\textbf{Category}}                                                                                                                                                           & \begin{tabular}[c]{@{}c@{}c@{}} \textbf{Methods}\end{tabular}               & \begin{tabular}[c]{@{}c@{}c@{}}\rotatebox{90}{Weighted graphs} \end{tabular} & \begin{tabular}[c]{@{}c@{}c@{}}\rotatebox{90}{Heterogeneous graphs\;\;}\end{tabular} & \begin{tabular}[c]{@{}c@{}c@{}}\rotatebox{90}{Attributed graphs}\end{tabular} & \begin{tabular}[c]{@{}c@{}c@{}}\rotatebox{90}{Feature propagation} \end{tabular} & \begin{tabular}[c]{@{}c@{}c@{}} \rotatebox{90}{Cross-graph interaction}\end{tabular} & \multicolumn{1}{c}{\textbf{Applications}}                                                                        \\ \toprule
\multirow{5}{*}{\begin{tabular}[c]{@{}c@{}}Graph Embedding \\ based GSL\end{tabular}}   & \multirow{3}{*}{\begin{tabular}[c]{@{}l@{}}Node-level Embedding\end{tabular}} & \cite{tixier2019graph}         & \xmark                                                         & \xmark                                                             & \cmark                                                         & \xmark                                                             & \xmark                                                              & \begin{tabular}[c]{@{}l@{}}Social Network Analysis\\ Bioinformatics \end{tabular}                        \\ \cline{3-9} 
                                                                                        &                                                                                                & \cite{nikolentzos2017matching} & \xmark                                                         & \cmark                                                            & \xmark                                                          & \xmark                                                             & \xmark                                                              & Chemoinformatics                                                                                      \\ \cline{2-9} 
                                                                                        & \begin{tabular}[c]{@{}l@{}}Graph-level Embedding\end{tabular}               & \cite{narayanan2017graph2vec,atamna2019spi,wu2018dgcnn}  & \xmark                                                         & \cmark                                                            & \xmark                                                          & \xmark                                                             & \xmark                                                              & \begin{tabular}[c]{@{}l@{}}Chemoinformatics\\ Bioinformatics \end{tabular}                        \\ \cline{3-9} 
                                                                                        &                                                                                                & \cite{anonymous2019_inductive}  & \xmark                                                         & \cmark                                                            & \xmark                                                          & \xmark                                                             & \xmark                                                              & \begin{tabular}[c]{@{}l@{}}Social Network Analysis\\ Chemoinformatics \end{tabular}                        \\ \cline{3-9} 
                                                                                        &                                                                                                & \cite{xu2017neural} & \xmark                                                         & \xmark                                                            & \cmark                                                          & \xmark                                                             & \xmark                                                              & Binary Code Similarity                                                                                      \\ \cline{3-9} 
                                                                                        &                                                                                                & \cite{liu2019n} & \xmark                                                         & \xmark                                                            & \cmark                                                          & \xmark                                                             & \xmark                                                              & Chemoinformatics                                                                                      \\ \hline
\multirow{7}{*}{GNN-based GSL}                                                          & \multirow{2}{*}{GNN-CNN Models}                                                          & \cite{bai2018convolutional}    & \xmark                                                         & \cmark                                                            & \cmark                                                         & \cmark                                                            & \xmark                                                              & Chemoinformatics                                                                                      \\ \cline{3-9} 
                                                                                        &                                                                                                & \cite{bai2019simgnn}           & \xmark                                                         & \cmark                                                            & \cmark                                                         & \cmark                                                            & \xmark                                                              & Chemoinformatics                                                                                      \\ \cline{2-9} 
                                                                                        & \multirow{2}{*}{Siamese GNNs}                                                                  & \cite{ktena2018metric,ma2019similarity,liu2019community}         & \cmark                                                        & \xmark                                                             & \cmark                                                         & \cmark                                                            & \xmark                                                              & Brain Network Analysis                                                                                    \\ \cline{3-9} 
                                                                                        % &                                                                                                & \cite{ma2019similarity}        & \cmark                                                        & \xmark                                                             & \cmark                                                         & \cmark                                                            & \xmark                                                              & Brain Network Analysis                                                                                    \\ \cline{3-9} 
                                                                                        &                                                                                                & \cite{ijcai2019-522}                            &    \xmark                                                        & \cmark                                                               & \cmark                                                            & \cmark                                                               & \xmark                                                                & Malware Detection                                                                                    \\ \cline{3-9} 
                                                                                      
                                                                                        &                                                                                                & \cite{chaudhuri2019siamese}                            &    \cmark                                                        & \xmark                                                               & \cmark                                                            & \cmark                                                               & \xmark                                                                & Image Retrieval                                                                                                                                                                 \\ \cline{2-9} 
                                                                                        & \multirow{2}{*}{\begin{tabular}[c]{@{}l@{}}GNN-based Graph\\ Matching Networks\end{tabular}}   & \cite{li2019graph,anonymous2019}             & \xmark                                                         & \xmark                                                             & \cmark                                                         & \cmark                                                            & \cmark                                                             & Binary Code Similarity                                                                                        \\ \cline{3-9} 
                                                                                        &                                                                                                & \cite{anonymous2019_mcs}             & \xmark                                                         & \cmark                                                             & \cmark                                                         & \cmark                                                            & \cmark                                                             & Chemoinformatics                                                                                        \\ \cline{3-9} 
                                                                                        &                                                                                                & \cite{wang2019learning,jiang2019glmnet}             & \xmark                                                         & \xmark                                                             & \cmark                                                         & \cmark                                                            & \cmark                                                             & Image Matching                                                                                        \\ \cline{3-9} 
                                                                                        &                                                                                                & \cite{guo2018neural}           & \cmark                                                        & \cmark                                                            & \cmark                                                         & \cmark                                                            & \xmark                                                              & 3D Action Recognition                                                                                     \\ \hline
\multirow{2}{*}{\begin{tabular}[c]{@{}c@{}}Deep Graph Kernel \\ based GSL\end{tabular}} & \begin{tabular}[c]{@{}l@{}}Sub-structure based \\ Deep Kernels\end{tabular}                    & \cite{yanardag2015deep}        & \xmark                                                         & \cmark                                                            & \xmark                                                          & \xmark                                                             & \xmark                                                              & \begin{tabular}[c]{@{}l@{}}Chemoinformatics\\ Bioinformatics \\ Social Network Analysis\end{tabular} \\ \cline{2-9} 
                                                                                        & \begin{tabular}[c]{@{}l@{}}Deep Neural Network \\ based Kernels\end{tabular}                   & \cite{al2019ddgk}              & \xmark                                                         & \cmark                                                            & \cmark                                                         & \xmark                                                             & \cmark                                                             & \begin{tabular}[c]{@{}l@{}}Chemoinformatics\\ Bioinformatics \end{tabular}                           \\ \cline{3-9} 
                                                                                        &                                                                                                & \cite{du2019graph}  & \xmark                                                         & \cmark                                                            & \cmark                                                          & \cmark                                                             & \xmark                                                              & \begin{tabular}[c]{@{}l@{}}Social Network Analysis\\ Chemoinformatics \end{tabular}                        \\ 
                                                                                        % \hline
\bottomrule
\end{tabular}
}
\end{center}
% \end{adjustbox}
\vspace{-4mm}
\end{table*}
}

% \textcolor{red}{\textbf{TODO:} This should be edited to be precise, and focus on the general categories, not the methods, and same should be done at the beginning of each category}

% With the rapid development of deep neural networks, strategies based on deep models have emerged as a new direction to approach the graph similarity learning problem. Some of the existing work in this area employ graph embedding techniques to learn node-level or graph-level representations from the input graphs and learn the similarity between graphs based on these representations \cite{tsitsulin2018verse,tixier2019graph}. Another large body of work are based on graph neural networks (GNN), which learns deep graph representation and leverage it for the similarity learning task \cite{bai2018convolutional,bai2019simgnn,ma2019similarity,li2019graph}. Another kind of deep strategy has also emerged which proposes to construct deep divergence graph kernels for capturing graph similarity \cite{al2019ddgk}. A taxonomy of these work is summarized in Table \ref{tab:taxonomy}. In this section, we will discuss on these different deep models. 
% In the area of deep graph similarity learning, there has been various approaches proposed for this learning task. Based on how the deep graph representation learning module is leveraged for the similarity learning,
\gm{In this section, we describe the taxonomy for the literature of deep graph similarity learning. As shown in Fig. ~\ref{fig:taxonomy}, we propose two intuitive taxonomies for categorizing the various deep graph similarity learning methods based on the model architecture and the type of features used in these methods.} 

\gm{First, we start by discussing the categorization based on which model architecture has been used. There are three main categories of deep graph similarity learning methods (see Fig.~\ref{fig:model_taxonomy}): (1) graph embedding based methods, which apply graph embedding techniques to obtain node-level or graph-level representations and further use the representations for similarity learning \cite{tixier2019graph,nikolentzos2017matching,narayanan2017graph2vec,atamna2019spi,wu2018dgcnn,anonymous2019_inductive,xu2017neural,liu2019n}; (2) graph neural network (GNN) based models, which are based on using GNNs for similarity learning, including GNN-CNNs \cite{bai2018convolutional,bai2019simgnn}, Siamese GNNs \cite{ktena2018metric,ma2019similarity,liu2019community,ijcai2019-522,chaudhuri2019siamese} and GNN-based graph matching networks \cite{li2019graph,anonymous2019,anonymous2019_mcs,wang2019learning,jiang2019glmnet,guo2018neural}; and (3) deep graph kernels that first map graphs into a new feature space, where kernel functions are defined for similarity learning on graph pairs, including sub-structure based deep kernels \cite{yanardag2015deep} and deep neural network based kernels \cite{al2019ddgk,du2019graph}. In the meantime, different methods may use different types of features in the learning process.}

\gm{Second, we discuss the categorization of methods based on the type of features used in them. Existing GSL approaches can be generally grouped into two categories (see Fig.~\ref{fig:feature_taxonomy}): (1) methods that uses single-graph features
~\cite{ktena2018metric,ma2019similarity,liu2019community,ijcai2019-522,chaudhuri2019siamese};
% ~\cite{tixier2019graph,nikolentzos2017matching,narayanan2017graph2vec,atamna2019spi,wu2018dgcnn,anonymous2019_inductive,xu2017neural,yanardag2015deep,bai2018convolutional,ktena2018metric,ma2019similarity,liu2019community,ijcai2019-522,chaudhuri2019siamese}; 
(2) methods that uses cross-graph features for similarity learning~\cite{li2019graph,anonymous2019,anonymous2019_mcs,al2019ddgk,wang2019learning,anonymous2019_mcs}. The main difference between these two categories of methods is that for methods using single-graph features, the representation of each graph is learned individually,  while those methods that use cross-graph features allow graphs to learn and propagate features from each other and the cross-graph interaction is leveraged for pairs of graphs. The single-graph features mainly includes graph embeddings at different granularity (i.e.,node-level, graph-level, and subgraph-level), while the cross-graph features includes the cross-graph node-level features and cross-graph graph-level features, which are usually obtained by node-level attention and graph-level attention across the two graphs in each pair.}

Next, we detail the description of the methods based on the taxonomy in Figures~\ref{fig:model_taxonomy} and~\ref{fig:feature_taxonomy}. We summarize the general characteristics and applications of all the methods in Table~\ref{tab:taxonomy}, including the type of graphs they are developed for, the type of features, and the domains/applications where they could be applied. We describe these methods in the following order:

\begin{enumerate}
    \item Graph embedding based GSL
    \item Graph Neural Network based GSL
    \item Deep graph kernel based GSL
\end{enumerate}
\noindent
 
%The graph embedding based GSL works first learn the representation for each graph using deep graph embedding strategies in a separate stage, after which the output embeddings are used for learning the similarity scores for pairs of graphs. The GNN-based GSL methods, instead, solve the similarity learning problem using GNNs in an end-to-end fashion. Usually a dot product layer or fully connected layers are added after the graph convolutional layers for producing or predicting the similarity scores between two graphs, and the representation learning via graph convolutions on each graph in a pair influences each other usually by weight sharing or cross-graph attention mechanism. The deep graph kernel based methods aim to define a kernel function that measures the similarity between graphs, where the key is to define a kernel that captures the semantic inherent in the graph representations.   

\subsection{Graph Embedding based Graph Similarity Learning}
% While the graph similarity work discussed in Section 4.1 - 4.3 calculate graph similarity scores with the predefined measures, in this section, we will introduce the category of methods that aim to learn the graph similarity scores with an end-to-end learning process.
Graph embedding has received considerable attention in the past decade \cite{cui2018survey,zhang2018network}, and a variety of deep graph embedding models have been proposed in recent years \cite{huang2019learning,narayanan2017graph2vec,gao2019graph1}, \gm{for example the popular \textit{DeepWalk} model proposed in \cite{perozzi2014deepwalk} and the \textit{node2vec} model from \cite{grover2016node2vec}}. Similarity learning methods based on graph embedding seek to utilize node-level or graph-level representations learned by these graph embedding techniques for defining similarity functions or predicting similarity scores \cite{tsitsulin2018verse,tixier2019graph,narayanan2017graph2vec}. Given a collection of graphs, these works first aim to convert each graph $G$ into a $d-$dimensional space $(d\ll\|V\|)$, where the graph is represented as either a set of $d-$dimensional vectors with each vector representing the embedding of one node (i.e.,node-level embedding) or a $d-$dimensional vector for the whole graph as the graph-level embedding \cite{cai2018comprehensive}. The graph embeddings are usually learned in an unsupervised manner in a separate stage prior to the similarity learning stage, where the graph embeddings obtained are used for estimating or predicting the similarity score between each pair of graphs.   

\subsubsection{Node-level Embedding based Methods}
Node-level embedding based methods compare graphs using the node-level representations learned from the graphs. The similarity scores obtained by these methods mainly capture the similarity between the corresponding nodes in two graphs. Therefore they focus on the local node-level information on graphs during the learning process.   

\paragraph{\textbf{node2vec-PCA}.} In \cite{tixier2019graph}, the node2vec approach \cite{grover2016node2vec} is employed for obtaining the node-level embeddings of graphs. To make the embeddings of all the graphs in the given collection comparable, they apply the principal component analysis (PCA) on the embeddings to retain the first $d \ll D$ principal components (where $D$ is the dimensionality of the original node embedding space). Afterwards, \gm{the embedding matrix of each graph is split into $d/2$ 2D slices. Suppose there are $n$ nodes in each graph $G$ and the embedding matrix for graph $G$ is $F \in \mathbb{R}^{n\times d}$, then $d/2$ 2D slices each with $\mathbb{R}^{n\times 2}$ will be obtained, which are viewed as $d/2$ \textit{channels}. Then each 2D slice from the embedding space is turned into regular grids by discretizing them into a fixed number of equallly-sized bins, where the value associate with each bin is the count of the number of nodes falling into that bin. These bins can be viewed as \textit{pixels}.} Then, the graph is represented as a stack of 2D histograms of its node embeddings. The graphs are then compared in the grid space and input into a 2D CNN as multi-channel image-like structures for a graph classification task. 

\paragraph{\textbf{Bag-of-Vectors}.} In \cite{nikolentzos2017matching}, the nodes of the graphs are first embedded in the Euclidean space using the eigenvectors of the adjacency matrices of the graphs, and each graph is then represented as a bag-of-vectors. The similarity between two graphs is then measured by computing a matching based on the Earth Mover's Distance \cite{rubner2000earth} between the two sets of embeddings. 

Although node embedding based graph similarity learning methods have been extensively developed, a common problem with these methods is that, since the comparison is based on node-level representations, the global structure of the graphs tends to be ignored, which actually is very important for comparing two graphs in terms of their structural patterns. 
%node representations are learned independently between graphs, they are generally less suitable for graph similarity computations compared to the methods that use measures or representations of the entire graph.
\subsubsection{Graph-level Embedding based Methods}
The graph-level embedding based methods aim to learn a vector representation for each graph and then learn the similarity score between graphs based on their vector representations. 

\paragraph{\textbf{(1) graph2vec.}} In \cite{narayanan2017graph2vec}, a graph2vec was proposed to learn distributed representations of graphs, similar to Doc2vec~\cite{le2014distributed} in natural language processing. In graph2vec each graph is viewed as a document and the rooted subgraphs around every node in the graph are viewed as words that compose the document. \gm{There are two main components in this method: first, a procedure to extract rooted subgraphs around every node in a given graph following the Weisfeiler-Lehman relabeling process and second, the procedure to learn embeddings of the given graphs by skip-gram with negative sampling. The Weisfeiler-Lehman relabeling algorithm takes the root node of the given graph and degree of the intended subgraph $d$ as inputs, and returns the intended subgraph.}
\gm{In the negative sampling phase, given a graph and a set of rooted subgraphs in its context, a set of randomly chosen subgraphs are selected as negative samples and only the embeddings of the negative samples are updated in the training.} After the graph embedding is obtained for each graph, the similarity or distance between graphs are computed in the embedding space for downstream prediction tasks (e.g., graph classification, clustering, etc.).

\paragraph{\textbf{(2) Neural Networks with Structure2vec.}} In \cite{xu2017neural}, a deep graph embedding approach is proposed for cross-platform binary code similarity detection. A Siamese architecture is applied to enable the pair-wise similarity learning, and the graph embedding network based on Structure2vec \cite{dai2016discriminative} is used for learning graph representations in the twin networks, which share weights with each other. \gm{The Structure2vec is a neural network approach inspired by graphical model inference algorithms where node-specific features are aggregated recursively according to graph topology. After a few steps of recursion, the network will produce a new feature representation for each node which considers both graph characteristics and long-range interaction between node features.} Given is a set of $K$ pairs of graphs $<G_i, {G_i}^\prime>$, with ground truth pair label $y_i \in \{+1,-1\}$, where $y_i = +1$ indicates that $G_i$ and ${G_i}^\prime$ are similar, and $y_i = -1$ indicates they are dissimilar. With the Structure2vec embedding output for $G_i$ and ${G_i}^\prime$, represented as $\mathbf{f}_i$ and ${\mathbf{f}_i}^\prime$ respectively, they define the Siamese network output for each pair as 
\begin{align}
    Sim(G_i,{G_i}^\prime) = \cos(\mathbf{f}_i,{\mathbf{f}_i}^\prime) = \frac{\langle \mathbf{f}_i,{\mathbf{f}_i}^\prime \rangle}{\|\mathbf{f}_i\| \cdot \| {\mathbf{f}_i}^\prime \|}
\end{align}
and the following loss function is used for training the model. 
\begin{align}
    L = \sum_{i=1}^{K} (Sim(G_i,{G_i}^\prime) - y_i)^2
    \label{eq:loss1}
\end{align}

\paragraph{\textbf{(3) Simple Permutation-Invariant GCN.}} In \cite{atamna2019spi}, a graph representation learning method based on a simple permutation-invariant graph convolutional network is proposed for the graph similarity and graph classification problem. A graph convolution module is used to encode local graph structure and node features, after which a sum-pooling layer is used to transform the substructure feature matrix computed by the graph convolutions into a single feature vector representation of the input graphs. The vector representation is then used as features for each graph, based on which the graph similarity or graph classification task can be performed. 

\paragraph{\textbf{(4) SEED: Sampling, Encoding, and Embedding Distributions.}} In \cite{anonymous2019_inductive}, an inductive and unsupervised graph representation learning approach called SEED is proposed for graph similarity learning. The proposed framework consists of three components: sampling, encoding, and embedding distribution. In the sampling stage, a number of subgraphs called WEAVE are sampled based on the random walk with earliest visit time. Then in the encoding stage, an autoencoder \cite{hinton2006reducing} is used to encode the subgraphs into dense low-dimensional vectors. Given a set of k sampled WEAVEs $\{X_1, X_2, X_3,\cdots,X_k\}$, for each subgraph $X_i$ the autoencoder works as follows. 
\begin{align}
    \mathbf{z}_i = f(X_i;{\theta}_e), \quad
    \hat{X_i} = g(\mathbf{z}_i;{\theta}_d),
\end{align}
\noindent where $\mathbf{z}_i$ is the dense low-dimensional representation for the input WEAVE subgraph $X_i$, $f(\cdot)$ is the encoding function implemented with an Multi-layer Perceptron (MLP) with parameters ${\theta}_e$, and $g(\cdot)$ is the decoding function implemented by another MLP with parameters ${\theta}_d$. A reconstruction loss is used to train the autoencoder:
\begin{align}
    L = \norm{X - \hat{X}}_2^2
\end{align}
\noindent After the autoencoder is well trained, the final subgraph embedding vectors ${\mathbf{z}_1,\mathbf{z}_2, \mathbf{z}_3,\cdots,}$ and $\mathbf{z}_k$ can be obtained for each graph. Finally, in the embedding distribution stage, the distance between the subgraph distributions of two input graphs $G$ and $H$ is evaluated using the maximum mean discrepancy (MMD) \cite{gretton2012kernel} on the embeddings. Assume the $k$ subgraphs sampled from $G$ are encoded into embeddings ${\mathbf{z}_1,\mathbf{z}_2, \cdots,\mathbf{z}_k}$, and the $k$ subgraphs of $H$ are encoded into embeddings ${\mathbf{h}_1,\mathbf{h}_2, \cdots,\mathbf{h}_k}$, the MMD distance between $G$ and $H$ is:
\begin{align}
    \widehat{\text{MMD}}(G,H) = \norm{\hat{\mu}_G - \hat{\mu}_H}_2^2
\end{align}
\noindent where $\hat{\mu}_G$ and $\hat{\mu}_H$ are empirical kernel embeddings of the two distributions, which are defined as: 
\begin{align}
    \hat{\mu}_G = \frac{1}{k}\sum_{i=1}^{k}\phi(\mathbf{z}_i), \quad \hat{\mu}_H = \frac{1}{k}\sum_{i=1}^{k}\phi(\mathbf{h}_i)
\end{align}
\noindent where $\phi(\cdot)$ is the feature mapping function used for the kernel function for graph similarity evaluation. An identity kernel is applied in this work.  

\paragraph{\textbf{(5) DGCNN: Disordered Graph CNN}.} In \cite{wu2018dgcnn}, another graph-level representation learning approach called DGCNN is introduced based on graph CNN and mixed Gaussian model, where a set of key nodes are selected from each graph. Specifically, to ensure the number of neighborhoods of the nodes in each graph is consistent, the same number of key nodes are sampled for each graph in a key node selection stage. Then a convolution operation is performed over the kernel parameter matrix and the nodes in the neighborhood of the selected key nodes, after which the graph CNN takes the output of the convolutional layer as the input data of the overall connection layer. Finally, the output of the dense hidden layer is used as the feature vector for each graph in the graph similarity retrieval task.

\paragraph{\textbf{(6) N-Gram Graph Embedding.}} In \cite{liu2019n}, an unsupervised graph representation based method called $N$-gram is proposed for similarity learning on molecule graphs. It first views each node in the graph as one token and applies an analog of the CBOW (continuous bag of words) \cite{mikolov2013distributed} strategy and trains a neural network to learn the node embeddings for each graph. Then it enumerates the walks of length $n$ in each graph, where each walk is called an $n$-gram, and obtains the embedding for each $n$-gram by assembling the embeddings of the nodes in the $n$-gram using element-wise product. \gm{The embedding for the n-gram walk set is defined as the sum of the embeddings for all n-grams. The final n-gram graph-level representation up to lenght $T$ is then constructed by concatenating the embeddings of all the $n$-gram sets for $n\in \{1,2,\cdots,T\}$ in the graph.} Finally, the graph-level embeddings are used for the similarity prediction or graph classification task for molecule analysis.     

\gm{By summarizing the embedding based methods, we find the main advantage of these methods is their speed and scalability, due to the fact that the graph representations learned through these factorized models are developed on each single graph where there is no feature interactions across graphs. This property makes these methods a great option for graph similarity learning applications such as graph retrieval, where similarity search becomes a nearest neighbor search in a database of the precomputed graph representations by these factorized methods.} Moreover, these embedding based methods provide a variety of perspectives and strategies for learning representations from graphs and demonstrate that these representations can be used for graph similarity learning. However, there are also shortcomings in these solutions, a common one being that the embeddings are learned independently on the individual graphs in a separate stage from the similarity learning, therefore the graph-graph proximity is not considered or utilized in the graph representation learning process, and the representations learned by these models may not be suitable for graph-graph similarity prediction compared to the methods that integrate the similarity learning with the graph representation learning in an end-to-end framework. 

\subsection{GNN-based Graph Similarity Learning}
\begin{figure}[t]
    \centering
    \includegraphics[width = 0.95 \columnwidth]{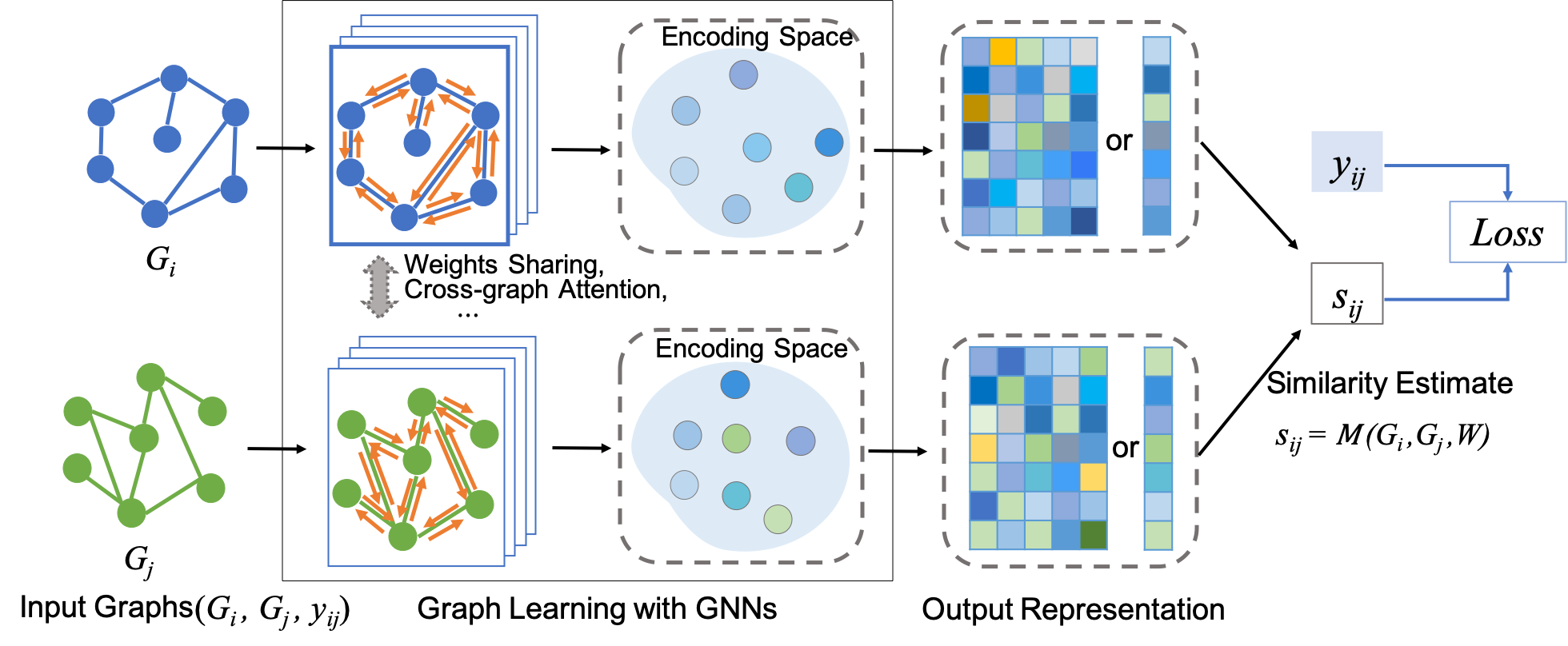}
    \caption{Illustration of GNN-based Graph Similarity Learning.}
    \label{fig:gsl}
\end{figure}

\gm{The similarity learning methods based on Graph Neural Networks (GNNs) seek to learn graph representations by GNNs while doing the similarity learning task in an end-to-end fashion. Fig.~\ref{fig:gsl} illustrates a general workflow of GNN-based graph similarity learning models. Given pairs of input graphs $<G_i, G_j, y_{ij}>$, where $y_{ij}$ denotes the ground-truth similarity label or score of $<G_i, G_j>$, the GNN-based GSL methods first employ multi-layer GNNs with weights $W$ to learn the representations for $G_i$ and $G_j$ in the encoding space, where the learning on each graph in a pair could influence each other by some mechanisms such as weight sharing and cross-graph interactions between the GNNs for the two graphs. A matrix or vector representation will be output for each graph by the GNN layers, after which a dot product layer or fully connected layers can be added to produce or predict the similarity scores between two graphs. Finally, the similarity estimates for all pairs of graphs and their ground-truth labels are used in a loss function for training the model $M$ with parameters $W$.}

Before introducing the methods in this category, we provide the necessary background on GNNs.

\paragraph{\textbf{GNN Preliminaries}.} Graph neural networks (GNNs) were first formulated in \cite{gori2005new}, which proposed to use a propagation process to learn node representations for graphs. It has then been further extended by \cite{scarselli2008graph,gallicchio2010graph}. Later, graph convolutional networks were proposed which compute node updates by aggregating information in local neighborhoods \cite{bruna2013spectral,defferrard2016convolutional,kipf2016semi}, and they have become the most popular graph neural networks, which are widely used and extended for graph representation learning in various domains \cite{zhou2018graph,zhang2018graph,gao2018large,gao2019graph,gao2019graph1}. 

With the development of graph neural networks, researchers began to build graph similarity learning models based on GNNs. In this section, we will first introduce the workflow of GCNs with the spectral GCN \cite{shuman2013emerging} as an example, and then describe the GNN-based graph similarity learning methods covering three main categories.

Given a graph $G=(V, E, \mathbf{A})$, where $V$ is the set of vertices, $E \subset V \times V $ is the set of edges, and  $\mathbf{A} \in \mathbb{R}^{m \times m}$ is the adjacency matrix, the diagonal degree matrix $\mathbf{D}$ will have elements $\mathbf{D}_{ii} = \sum_j \mathbf{A}_{ij}$. The graph Laplacian matrix is $\mathbf{L} = \mathbf{D} - \mathbf{A}$, which can be normalized as $\mathbf{L} = \mathbf{I}_m - \mathbf{D}^{-\frac{1}{2}}\mathbf{A} \mathbf{D}^{-\frac{1}{2}}$, where $\mathbf{I}_m$ is the identity matrix. Assume the orthonormal eigenvectors of $\mathbf{L}$ are represented as $\{u_l\}_{l=0}^{m-1}\in \mathbb{R}^{m \times m}$, and their associated eigenvalues are $\{\lambda_l\}_{l=0}^{m-1}$, the Laplacian is diagonalized by the Fourier basis $[u_0, \cdots,u_{m-1}](=\mathbf{U})\in \mathbb{R}^{m \times m}$ and $\mathbf{L} = \mathbf{U\Lambda U^T}$ where $\mathbf{\Lambda} = diag([\lambda_0,\cdots,\lambda_{m-1}])\in \mathbb{R}^{m\times m}$. The graph Fourier transform of a signal $x\in \mathbb{R}^m$ can then be defined as $\hat{x} = \mathbf{U^T}x \in \mathbb{R}^m$\cite{shuman2013emerging}. Suppose a signal vector $\mathbf{x} : V \rightarrow \mathbb{R}$ is defined on the nodes of graph $G$, where $\mathbf{x}_i$ is the value of $\mathbf{x}$ at the $i^{th}$ node. Then the signal $\mathbf{x}$ can be filtered by $g_\theta$ as
\vspace{-2mm}
\begin{align}
    y = g_\theta*\mathbf{x} = g_\theta(\mathbf{L})\mathbf{x} = g_\theta(\mathbf{U{\Lambda}U^T})\mathbf{x} = \mathbf{U}g_\theta(\Lambda)\mathbf{U^T}\mathbf{x}
\label{eq:filter_signal}
\end{align}
where the filter $g_\theta(\Lambda)$ can be defined as $g_{\theta}(\Lambda) = \sum_{k=0}^{K-1}{\theta_k}{\Lambda^k}$, and the parameter $\theta\in {\mathbb{R}}^K$ is a vector of polynomial coefficients \cite{defferrard2016convolutional}. GCNs can be constructed by stacking multiple convolutional layers in the form of Equation (\ref{eq:filter_signal}), with a non-linearity activation (ReLU) following each layer.

%Some of these work use graph neural networks to learn graph representations and use the learned representations for predicting similarity scores, which is approached as a regression problem. Some other work use the Siamese network architecture with GNNs as twin networks to simultaneously learn from two graphs and the graph similarity estimate is usually leveraged in a contrastive loss function. The rest works consider some matching mechanisms during the learning with GNNs, and the graph-graph proximity is explicitly considered in the graph representation process. 

% \textcolor{red}{\textbf{TODO:} Add descriptions for Figure 1.}

Based on how graph-graph similarity/proximity is leveraged in the learning, we summarize the existing GNN-based graph similarity learning work into three main categories: 1) GNN-CNN mixed models for graph similarity prediction, 2) Siamese GNNs for graph similarity prediction, and 3) GNN-based graph matching networks.

\subsubsection{GNN-CNN Models for Graph Similarity Prediction} 
The works that use GNN-CNN mixed networks for graph similarity prediction mainly employ GNNs to learn graph representations and leverage the learned representations into CNNs for predicting similarity scores, which is approached as a classification or regression problem. Fully connected layers are often added for the similarity score prediction in an end-to-end learning framework. 

\paragraph{\textbf{(1) GSimCNN.}} In \cite{bai2018convolutional}, a method called GSimCNN is proposed for pairwise graph similarity prediction, which consists of three stages. In Stage 1, node representations are first generated by multi-layer GCNs, where each layer is defined as 
\begin{align}
    conv(\mathbf{x}_i) = ReLU(\sum_{j \in N(i)}\frac{1}{\sqrt{d_id_j}}\mathbf{x}_j\mathbf{W}^{(l)} + \mathbf{b}^{(l)})
\end{align}

\noindent where $N(i)$ is the set of first-order neighbors of node $i$ plus node $i$ itself, $d_i$ is the degree of node $i$ plus $1$, $\mathbf{W}^{(l)}$ is the weight matrix for the $l-$th GCN layer, $\mathbf{b}^{(l)}$ is the bias, and $ReLU(x) = max(0,x)$ is the activation function. In Stage 2, the inner products between all possible pairs of node embeddings between two graphs from different GCN layers are calculated, which results in multiple similarity matrices. Finally, the similarity matrices from different layers are processed by multiple independent CNNs, where the output of the CNNs are concatenated and fed into fully connected layers for predicting the final similarity score $s_{ij}$ for each pair of graphs $G_i$ and $G_j$.

\paragraph{\textbf{(2) SimGNN.}}In \cite{bai2019simgnn}, a SimGNN model is introduced based on the GSimCNN from \cite{bai2018convolutional}. In addition to pairwise node comparison with node-level embeddings from the GCN output, neural tensor networks (NTN) \cite{socher2013reasoning} are utilized to model the relation between the graph-level embeddings of two input graphs, whereas the graph embedding for each graph is generated via a weighted sum of node embeddings, and a global context-aware attention is applied on each node, such that nodes similar to the global context receive higher attention weights. Finally, both the comparison between node-level embeddings and graph-level embeddings are considered for the similarity score prediction in the CNN fully connected layers.

% learnable embedding function is designed to map every graph into an embedding vector based on GNN, which provides a global summary of a graph. An attention mechanism is proposed to emphasize the important nodes with respect to a specific similarity metric. Then a pairwise node comparison method is designed to supplement the graph-level embeddings with fine-grained node-level information.  

%In \cite{li2019graph}, a GNN embedding model is proposed, which consists of three parts: an encoder, propagation layers, and an aggregator. The encoder maps the node and edge features to initial node and edge vectors via separate MLPs. Propagation layers maps the node vectors to new representations by accumulating information in local neighborhood. Then after a few rounds of propagations, an aggregator is applied for computing a graph-level representation from the node representations. Finally, the similarity between two graphs are computed in the vector space of the graph-level representations. Pairwise loss and triplet loss are used separately for training the model. In \cite{baiunsupervised}, a graph-level embedding learning model is proposed which learns graph-level representations for graphs that can be used for approximating graph-graph proximity. Graph Isomophism Network \cite{xu2018powerful} is used for obtaining node embeddings, and a Multi-Scale Node Attention (MSNA) is proposed for transforming node embeddings into graph-level embeddings. Graph edit distance is used to measure the graph-graph proximity.

\subsubsection{Siamese GNN models for Graph Similarity Learning} 
\begin{figure}[t]
    \centering
    \includegraphics[width = 0.45 \columnwidth]{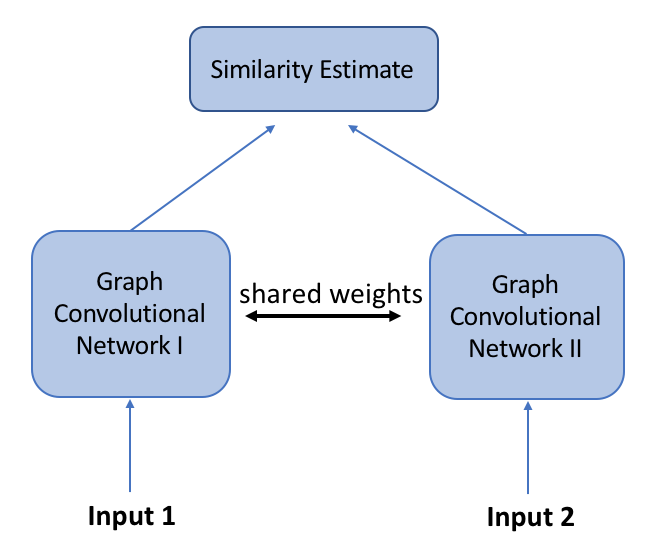}
    \caption{Siamese Architecture with Graph Convolutional Networks.}
    \label{fig:siamese}
\end{figure}

This category of works uses the Siamese network architecture with GNNs as twin networks to simultaneously learn representations from two graphs, and then obtain a similarity estimate based on the output representations of the GNNs. Fig.~\ref{fig:siamese} shows an example of Siamese architecture with GCNs in the twin networks, where the weights of the networks are shared with each other. The similarity estimate is typically leveraged in a loss function for training the network. 

% Siamese network was first introduced in \cite{{bromley1994signature}} for the signature verification problem, and it has been shown to be an effective architecture for solving image matching problems such as image recognition \cite{chopra2005learning}. Recently, the Siamese architecture has been introduced to the graph domain and has shown its advantage with GNNs for the graph similarity learning task \cite{ktena2018metric,ma2019similarity,ijcai2019-522}. 

%As the parameters between the twin neural networks in the Siamese network are shared, each input will be processed in the same way respectively in the twin networks, which can guarantee that similar input samples not be mapped to very different locations by the respective networks.

\paragraph{\textbf{(1) Siamese GCN}.} The work in~\cite{ktena2018metric} proposes to learn a graph similarity metric using the Siamese graph convolutional neural network (S-GCN) in a supervised setting. The S-GCN takes a pair of graphs as inputs and employs spectral GCN to get graph embedding for each input graph, after which a dot product layer followed by a fully connected layer is used to produce the similarity estimate between the two graphs in the spectral domain. 

\paragraph{\textbf{(2) Higher-order Siamese GCN.}} Higher-order Siamese GCN (HS-GCN) is proposed in \cite{ma2019similarity}, which incorporates higher-order node-level proximity into graph convolutional networks so as to perform higher-order convolutions on each of the input graphs for the graph similarity learning task. A Siamese framework is employed with the proposed higher-order GCN in each of the twin networks. Specifically, random walk is used for capturing higher-order proximity from graphs and refining the graph representations used in graph convolutions. Both this work and the S-GCN \cite{ktena2018metric} introduced above use the Hinge loss for training the Siamese similarity learning models: 
\begin{align}
    L_{Hinge} = \frac{1}{K}\sum_{i=1}^{N}\sum_{j=i+1}^{N} max(0,1-{y_{ij}}{s_{ij}}),
    \label{hinge_loss}
\end{align}
where $N$ is the total number of graphs in the training set, $K = N(N-1)/2$ is the total number of pairs from the training set, $y_{ij}$ is the ground-truth label for the pair of graphs $G_i$ and $G_j$ where $y_{ij} = 1$ for similar pairs and $y_{ij} = -1$ for dissimilar pairs, and $s_{ij}$ is the similarity score estimated by the model. More general forms of higher-order information (e.g., motifs~\cite{ahmed2015efficient,ahmed2017graphlet}) have been used for learning graph representations~\cite{rossi2018higher,rossi2020structural} and would likely benefit the learning.

\paragraph{{\textbf{(3) Community-preserving Siamese GCN}.}} In \cite{liu2019community}, another Siamese GCN based model called SCP-GCN is proposed for the similarity learning in functional and structural joint analysis of brain networks, where the graph structure used in the GCN is defined from the structural connectivity network while the node features come from the functional brain network. The contrastive loss (Equation~(\ref{eq:contrastive})) along with a newly proposed community-preserving loss (Equation~(\ref{eq:community_loss})) is used for training the model.

\begin{equation}
    L_{Contrastive} = \frac{y_{ij}}{2}\|\mathbf{g}_i - \mathbf{g}_j\|_2^2 + (1 - y_{ij})\frac{1}{2}\{max(0, m- \| \mathbf{g}_i - \mathbf{g}_j\|_2)\}^2
\label{eq:contrastive}    
\end{equation}

\noindent where $\mathbf{g}_i$ and $\mathbf{g}_j$ are the graph embeddings of graph $G_i$ and graph $G_j$ computed from the GCN, $m$ is a margin value which is greater than $0$. $y_{ij}=1$ if $G_i$ and $G_j$ are from the same class and $y_{ij}=0$ if they are from different classes. By minimizing the contrastive loss, the Euclidean distance between two graph embedding vectors will be minimized when the two graphs are from the same class, and maximized when they belong to different classes. The community-preserving loss is defined as follows.

\begin{equation} 
\label{eq:community_loss}
    L_{CP} = \alpha (\sum_{c}\frac{1}{|S_{c}|}\sum_{i \in S_{c}} \| \mathbf{z}_i - \hat{\mathbf{z}}_c\|_2^2) - \beta \sum_{c, c'} \| \hat{\mathbf{z}}_c - \hat{\mathbf{z}}_{c'} \|_2^2
\end{equation} 
where $S_c$ contains the indexes of nodes belonging to community $c$, $\hat{\mathbf{z}}_c = \frac{1}{|S_c|}\sum_{i \in S_c}\mathbf{z}_i$ is the community center embedding for each community $c$, where $\mathbf{z}_i$ is the embedding of the $i^{th}$ node, i.e., the $i^{th}$ row in the node embedding $\mathbf{Z}$ of the GCN output, and $\alpha$ and $\beta$ are the weights balancing the intra/inter-community loss.
%The first part of this loss computes the Euclidean distance between node embedding $\mathbf{z}_i$ and its community center $\hat{\mathbf{z}}_c$, while the second part computes the distance between community center $\hat{\mathbf{z}}_c$ and $\hat{\mathbf{z}}_{c'}$. 

\paragraph{\textbf{(4) Hierarchical Siamese GNN.}} In \cite{ijcai2019-522}, a Siamese network with two hierarchical GNN models is introduced for the similarity learning of heterogeneous graphs for unknown malware detection. Specifically, they consider the path-relevant sets of neighbors according to meta-paths and generate node embeddings by selectively aggregating the entities in each path-relevant neighbor set. The loss function in Equation~(\ref{eq:loss1}) is used for training the model. 

\paragraph{\textbf{(5) Siamese GCN for Image Retrieval.}} In \cite{chaudhuri2019siamese}, Siamese GCNs are used for content based remote sensing image retrieval, where each image is converted to a region adjacency graph in which each node represents a region segmented from the image. The goal is to learn an embedding space that pulls semantically coherent images closer while pushing dissimilar samples far apart. Contrastive loss is used in the model training.   

Since the twin GNNs in the Siamese network share the same weights, an advantage of the Siamese GNN models is that the two input graphs are guaranteed to be processed in the same manner by the networks. As such, similar input graphs would be embedded similarly in the latent space. Therefore, the Siamese GNNs are good for differentiating the two input graphs in the latent space or measuring the similarity between them. 

In addition to choosing the appropriate GNN models in the twin networks, one needs to choose a proper loss function. Another widely used loss function for Siamese network is the triplet loss \cite{schroff2015facenet}. For a triplet $(G_i, G_p, G_n)$, $G_p$ is from the same class as $G_i$, while $G_n$ is from a different class from $G_i$. The triplet loss is defined as follows. 
% \begin{equation}
%     L_{Contrastive} = \frac{1}{K}\sum_{i=1}^{N}\sum_{j=i+1}^{N}(\frac{y_{ij}}{2}d_{ij}^2 + (1 - y_{ij})\frac{1}{2}\{max(0, m - d_{ij})\}^2)
% \label{eq:loos_contrastive}    
% \end{equation}
% \noindent where $N$ is the total number of graphs in the training set, $K = N(N-1)/2$ is the total number of pairs from the training set, $y_{ij}$ is the ground-truth label for the pair of graphs $G_i$ and $G_j$ where $y_{ij} = 1$ for similar pairs and $y_{ij} = -1$ for dissimilar pairs, $d_{ij}$ represents the distance between $G_i$ and $G_j$ in the latent space, and $m$ is a margin value which is greater than 0. Having a margin indicates that dissimilar pairs that are beyond this margin will not contribute to the loss.

\begin{equation}
    L_{Triplet} = \frac{1}{K}\sum_K max(d_{ip} - d_{in} + m, 0)
    \label{eq:triplet_loss}
\end{equation}
\noindent where $K$ is the number of triplets used in the training, $d_{ip}$ represents the distance between $G_i$ and $G_p$, $d_{in}$ represents the distance between $G_i$ and $G_n$, and $m$ is a margin value which is greater than 0. By minimizing the triplet loss, the distance between graphs from same class (i.e., $d_{ip}$) will be pushed to $0$, and the distance between graphs from different classes (i.e.,$d_{in}$ will be pushed to be greater than $d_{ip} + m$. 

It is important to consider which loss function would be suitable for the targeted problem when applying these Siamese GNN models for the graph similarity learning task in practice.

\subsubsection{GNN-based Graph Matching Networks}
\begin{figure}[t]
\centering
    \begin{subfigure}[Siamese GNN]{		\includegraphics[width=.47\linewidth] {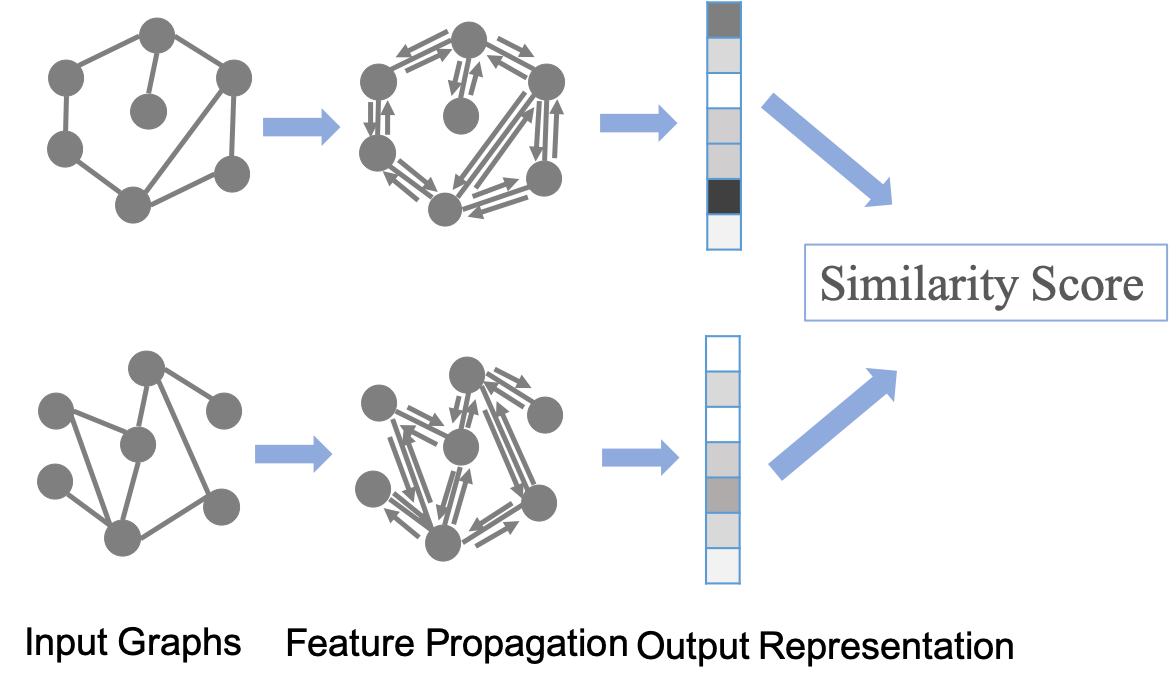}
		    \label{fig:sgnn}
		}%
		\end{subfigure}
		\begin{subfigure}[GNN-based Graph Matching Network]{
		\includegraphics[width=.47\linewidth]{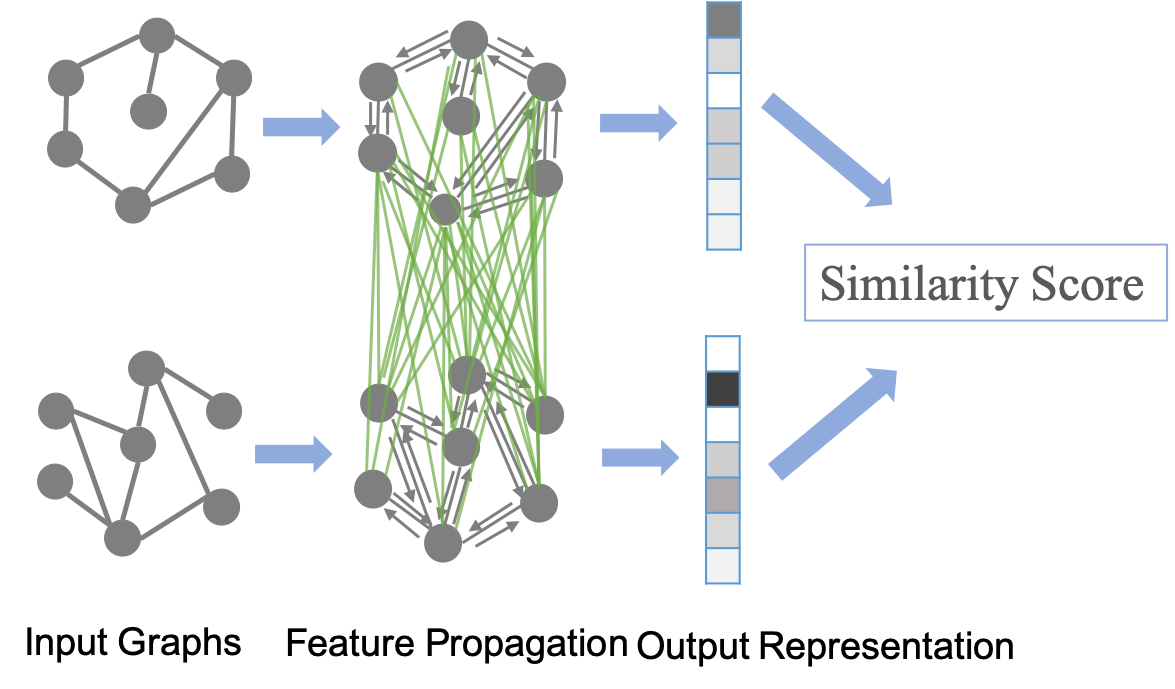}
		\label{fig:gmn}
		}%
		\end{subfigure}  
  \caption{Comparison of the Learning Process of Siamese GNN and GNN-based Graph Matching Network}
  \label{fig:sgnn_gmn}
\end{figure}
The work in this category adapts Siamese GNNs by incorporating matching mechanisms during the learning with GNNs, and cross-graph interactions are considered in the graph representation learning process. Fig.~\ref{fig:sgnn_gmn} shows this difference between the Siamese GNNs and the GNN-based graph matching networks.

\paragraph{\textbf{(1) GMN: Graph Matching Network.}} In \cite{li2019graph}, \gm{a GNN based architecture called Graph Matching Network (GMN) is proposed}, where the node update module in each propagation layer takes into account both the aggregated messages on the edges for each graph and a cross-graph matching vector which measures how well a node in one graph can be matched to the nodes in the other graph. Given a pair of graphs as input, the GMN jointly learns graph representations for the pair through the cross-graph attention-based matching mechanism, which propagates node representations by using both the neighborhood information within the same graph and cross-graph node information. A similarity score between the two input graphs is computed in the latent vector space. 
%Fig.~\ref{fig:gmn} shows the comparison of the GNN-based graph similarity learning without cross-graph matching, which are introduced in the above categories, and the proposed GMN with cross-graph attention-based matching.

\paragraph{\textbf{(2) NeuralMCS: Neural Maximum Common Subgraph GMN}.}Based on the graph matching network in \cite{li2019graph}, \cite{anonymous2019_mcs} proposes a neural maximum common subgraph (MCS) detection approach for learning graph similarity. The graph matching network is adapted to learn node representations for two input graphs $G_1$ and $G_2$, after which a likelihood of matching each node in $G_1$ to each node in $G_2$ is computed by a normalized dot product between the node embeddings. The likelihood indicates which node pair is most likely to be in the MCS, and the likelihood for all pairs of nodes constitutes the matching matrix $\mathbf{Y}$ for $G_1$ and $G_2$. Then a guided subgraph extraction process is applied, which starts by finding the most likely pair and iteratively expands the extracted subgraphs by selecting one more pair at a time until adding more pairs would lead to non-isomorphic subgraphs. To check the subgraph isomorphism, subgraph-level embeddings are computed by aggregating the node embeddings of the neighboring nodes that are included in the MCS, and Euclidean distance between the subgraph embeddings are computed. Finally, a similarity/match score is obtained based on the subgraphs extracted from $G_1$ and $G_2$.

\paragraph{\textbf{(3) Hierarchical Graph Matching Network.}} In \cite{anonymous2019}, a hierarchical graph matching network is proposed for graph similarity learning, which consists of a Siamese GNN for learning global-level interactions between two graphs and a multi-perspective node-graph matching network for learning the cross-level node-graph interactions between parts of one graph and one whole graph. Given two graphs $G_1$ and $G_2$ as inputs, a three-layer GCN is utilized to generate embeddings for them, and aggregation layers are added to generate the graph embedding vector for each graph. In particular, cross-graph attention coefficients are calculated between each node in $G_1$ and all the nodes in $G_2$, and between each node in $G_2$ and all the nodes in $G_1$. Then the attentive graph-level embeddings are generated using the weighted average of node embeddings of the other graph, and a multi-perspective matching function is defined to compare the node embeddings of one graph with the attentive graph-level embeddings of the other graph. Finally, the BiLSTM model \cite{schuster1997bidirectional} is used to aggregate the cross-level interaction feature matrix from the node-graph matching layer, followed by the final prediction layers for the similarity score learning. 

\paragraph{\textbf{(4) NCMN: Neural Graph Matching Network.}} In \cite{guo2018neural}, a Neural Graph Matching Network (NGMN) is proposed for few-shot 3D action recognition, where 3D data are represented as interaction graphs. A GCN is applied for updating node features in the graphs and an MLP is employed for updating the edge strength. A graph matching metric is then defined based on both node matching features and edge matching features. In the proposed NGMN, edge generation and graph matching metric are learned jointly for the few-shot learning task. 

Recently, deep graph matching networks were introduced for the graph matching problem for image matching \cite{anonymous2019_matching,zanfir2018deep,jiang2019glmnet,wang2019learning}. Graph matching aims to find node correspondence between graphs, such that the corresponding node and edge's affinity is maximized. Although the problem of graph matching is different from the graph similarity learning problem we focus on in this survey and is beyond the scope of this survey, some work on deep graph matching networks involves graph similarity learning and thus we review some of this work below to provide some insights into how deep similarity learning may be leveraged for graph matching applications, such as image matching.

\paragraph{\textbf{(5) GMNs for Image Matching.}}In \cite{jiang2019glmnet}, a Graph Learning-Matching Network is proposed for image matching. A CNN is first utilized to extract feature descriptors of all feature points for the input images, and graphs are then constructed based on the features. Then the GCNs are used for learning node embeddings from the graphs, in which both intra-graph convolutions and cross-graph convolutions are conducted. The final matching prediction is formulated as node-to-node affinity metric learning in the embedding space, and the constraint regularized loss along with cross-entropy loss is used for the metric learning and the matching prediction. In \cite{wang2019learning}, another GNN-based graph matching network is proposed for the image matching problem, which consists of a CNN image feature extractor, a GNN-based graph embedding component, an affinity metric function and a permutation prediction component, as an end-to-end learnable framework. Specifically, GCNs are used to learn node-wise embeddings for intra-graph affinity, where a cross-graph aggregation step is introduced to aggregate features of nodes in the other graph for incorporating cross-graph affinity into the node embeddings. The node embeddings are then used for building an affinity matrix which contains the similarity scores at the node level between two graphs, and the affinity matrix is further used for the matching prediction. The cross-entropy loss is used to train the model end-to-end.

\subsection{Deep Graph Kernels}

% \textcolor{red}{\textbf{TODO:} There are only two methods in the deep graph kernels, that's weird, that should be the most popular approach, please add a few more}
Graph kernels have become a standard tool for capturing the similarity between graphs for tasks such as graph classification \cite{vishwanathan2010graph}. Given a collection of graphs, possibly with node or edge attributes, the work in graph kernel aim to learn a kernel function that can capture the similarity between any two graphs. Traditional graph kernels, such as random walk kernels, subtree kernels, and shortest-path kernels have been widely used in the graph classification task \cite{giannis2019}. Recently, deep graph kernel models have also emerged, which build kernels based on the graph representations learned via deep neural networks.

\subsubsection{Deep Graph Kernels} 
In \cite{yanardag2015deep}, a Deep Graph Kernel approach is proposed. For a given set of graphs, each graph is decomposed into its sub-structures. Then the sub-structures are viewed as words and neural language models in the form of CBOW (continuous bag-of-words) and Skip-gram are used to learn latent representations of sub-structures from the graphs, where corpora are generated for the Shortest-path graph and Weisfeiler-Lehman kernels in order to measure the co-occurrence relationship between substructures. Finally, the kernel between two graphs is defined based on the similarity of the sub-structure space. 
\begin{figure}[t]
    \centering
    \includegraphics[width = 0.8 \columnwidth]{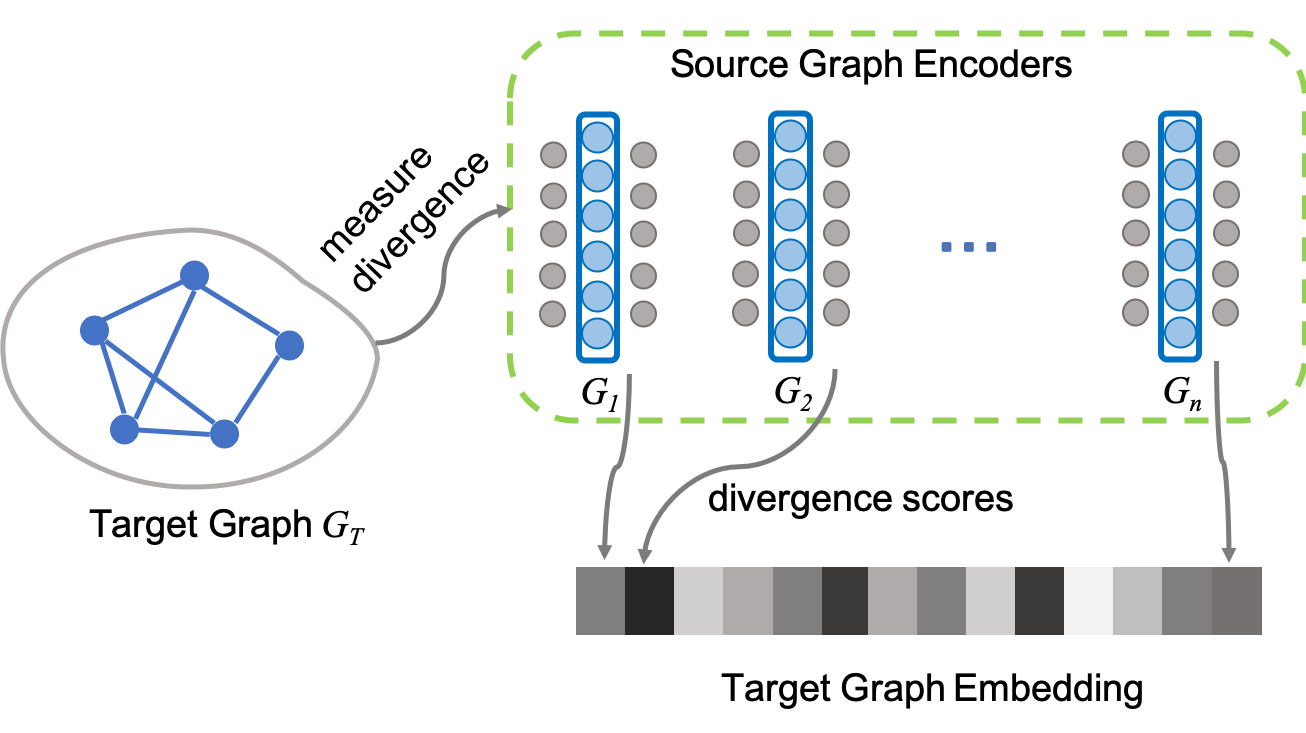}
    \caption{The Graph Representation Learning in the Deep Divergence Graph Kernels \cite{al2019ddgk}.}
    \label{fig:ddgk}
\end{figure}

\subsubsection{Deep Divergence Graph Kernels} In \cite{al2019ddgk}, a model called Deep Divergence Graph Kernels (DDGK) is introduced to learn kernel functions for graph pairs. Given two graphs $G_1$ and $G_2$, they aim to learn an embedding based kernel function $k( )$ as a similarity metric for graph pairs, defined as:
\begin{align}
    k(G_1,G_2) = \|\Psi(G_1) - \Psi(G_2)\|^2
\end{align}
\noindent where $\Psi(G_i)$ is a representation learned for $G_i$. This work proposes to learn graph representation by measuring the divergence of the target graph across a population of source graph encoders. Given a source graph collection $\{G_1, G_2,$ $\cdots, G_n\}$, a graph encoder is first trained to learn the structure of each graph in the source collection. Then, for a target graph $G_T$, the divergence of $G_T$ from each source graph is measured, after which the divergence scores are used to compose the vector representation of the target graph $G_T$. Fig.~\ref{fig:ddgk} illustrates the above graph representation learning process. Specifically, the divergence score between a target graph $G_T=(V_T,E_T)$ and a source graph $G_S=(V_S,E_S)$ is computed as follows:
\begin{align}
    \mathcal{D}^\prime (G_T \| G_S) = \sum_{v_i \in V_T} \sum_{\substack{j\\{e_{ij}\in E_T}}} -log \text{Pr}(v_j|v_i, H_S)
\end{align}
\noindent where $H_S$ is the encoder trained on graph $S$.

% The structure of the source graph is learned by passing it through a graph encoder. Then the source graph encoder is used to predict the structure of the target graph, so as to measure how much the target graph diverges from the source graph. Specifically, a cross-graph attention mechanism is proposed to learn a soft alignment between graphs. They use the attention-augmented encoder's predictions to define a divergence score for each pair of graphs, where the divergence is expected to be low if two graphs are similar. 

\subsubsection{Graph Neural Tangent Kernel} In \cite{du2019graph}, a Graph Neural Tangent Kernel (GNTK) is proposed for fusing GNNs with the neural tangent kernel, which is originally formulated for fully-connected neural networks in \cite{jacot2018neural} and later introduced to CNNs in \cite{arora2019exact}. Given a pair of graphs $<G,G^\prime>$, they first apply GNNs on the graphs. Let $f(\theta, G) \in \mathbb{R}$ be the output of the GNN under parameters $\theta \in \mathbb{R}^m$ on input Graph $G$, \gm{where $m$ is the dimension of the parameters}. To get the corresponding GNTK value, they calculate the expected value of
\begin{align}
    \Bigg \langle \frac{\partial f(\theta, G)}{\partial \theta}, \frac{\partial f(\theta, G^\prime )}{\partial \theta} \bigg \rangle
\end{align}
in the limit that $m \rightarrow \infty$ and $\theta$ are all Gaussian random variables.

Meanwhile, there are also some deep graph kernels proposed for the node representation learning on graphs for node classification and node similarity learning. For instance, in \cite{tian2019rethinking}, a learnable kernel-based framework is proposed for node classification, where the kernel function is decoupled into a feature mapping function and a base kernel. An encoder-decoder function is introduced to project each node into the embedding space and reconstructs pairwise similarity measurements from the node embeddings. Since we focus on the similarity learning between graphs in this survey, we will not discuss this work further.   

\section{Datasets and Evaluation}
% \usepackage{multirow}
\begin{table*}[]
\begin{center}
\caption{Summary of Benchmark Datasets that are Frequently Used in Deep Graph Similarity Learning.}
\label{tab:datasets}
\scalebox{0.78}{
\begin{tabular}{lclccccl}
\hline
\multicolumn{1}{|c|}{\textbf{Graph Type}}                          & \multicolumn{1}{c|}{\textbf{Datasets}}         & \multicolumn{1}{c|}{\textbf{Source}}                                           & \multicolumn{1}{c|}{\rotatebox{90}{Number of Graphs}} & \multicolumn{1}{c|}{\rotatebox{90}{\;\;Number of Classes\;\;}} & \multicolumn{1}{c|}{\rotatebox{90}{\;\;Avg. Number of Nodes\;\;}} & \multicolumn{1}{c|}{\rotatebox{90}{\;\;Avg. Number of Edges\;\;}} & \multicolumn{1}{c|}{\textbf{References}} \\ \hline
\multicolumn{1}{|c|}{\multirow{6}{*}{Social Networks}}  & \multicolumn{1}{c|}{COLLAB}           & \multicolumn{1}{c|}{\cite{yanardag2015deep}}         & \multicolumn{1}{c|}{5000}           & \multicolumn{1}{c|}{3}               & \multicolumn{1}{c|}{74.49}                & \multicolumn{1}{c|}{2457.78}              & \multicolumn{1}{c|}{\cite{yanardag2015deep,tixier2019graph,wu2018dgcnn,anonymous2019_inductive,du2019graph}}                            \\ \cline{2-8} 
\multicolumn{1}{|l|}{}                                  & \multicolumn{1}{c|}{IMDB-BINARY}      & \multicolumn{1}{c|}{\cite{yanardag2015deep}}         & \multicolumn{1}{c|}{1000}           & \multicolumn{1}{c|}{2}               & \multicolumn{1}{c|}{19.77}                & \multicolumn{1}{c|}{96.53}                & \multicolumn{1}{c|}{\cite{yanardag2015deep,atamna2019spi,anonymous2019_inductive,du2019graph}}                            \\ \cline{2-8} 
\multicolumn{1}{|l|}{}                                  & \multicolumn{1}{c|}{IMDB-MULTI}       & \multicolumn{1}{c|}{\cite{yanardag2015deep}}         & \multicolumn{1}{c|}{1500}           & \multicolumn{1}{c|}{3}               & \multicolumn{1}{c|}{13.00}                & \multicolumn{1}{c|}{65.94}                & \multicolumn{1}{c|}{\cite{yanardag2015deep,atamna2019spi,anonymous2019_inductive,bai2018convolutional,bai2019simgnn,du2019graph}}                            \\ \cline{2-8} 
\multicolumn{1}{|l|}{}                                  & \multicolumn{1}{c|}{REDDIT-BINARY}    & \multicolumn{1}{c|}{\cite{yanardag2015deep}}         & \multicolumn{1}{c|}{2000}           & \multicolumn{1}{c|}{2}               & \multicolumn{1}{c|}{429.63}               & \multicolumn{1}{c|}{497.75}               & \multicolumn{1}{c|}{\cite{yanardag2015deep,tixier2019graph}}                            \\ \cline{2-8} 
\multicolumn{1}{|l|}{}                                  & \multicolumn{1}{c|}{REDDIT-MULTI-5K}  & \multicolumn{1}{c|}{\cite{yanardag2015deep}}         & \multicolumn{1}{c|}{4999}           & \multicolumn{1}{c|}{5}               & \multicolumn{1}{c|}{508.52}               & \multicolumn{1}{c|}{594.87}               & \multicolumn{1}{c|}{\cite{yanardag2015deep,tixier2019graph}}                            \\ \cline{2-8} 
\multicolumn{1}{|l|}{}                                  & \multicolumn{1}{c|}{REDDIT-MULTI-12K} & \multicolumn{1}{c|}{\cite{yanardag2015deep}}         & \multicolumn{1}{c|}{11929}          & \multicolumn{1}{c|}{11}              & \multicolumn{1}{c|}{391.41}               & \multicolumn{1}{c|}{456.89}               & \multicolumn{1}{c|}{\cite{yanardag2015deep,tixier2019graph}}                            \\ \hline
\multicolumn{1}{|c|}{\multirow{3}{*}{Bioinformatics}}   & \multicolumn{1}{c|}{D\&D}             & \multicolumn{1}{c|}{\cite{dobson2003distinguishing}} & \multicolumn{1}{c|}{1178}           & \multicolumn{1}{c|}{2}               & \multicolumn{1}{c|}{284.32}               & \multicolumn{1}{c|}{715.66}               & \multicolumn{1}{c|}{\cite{al2019ddgk,nikolentzos2017matching,wu2018dgcnn}}                            \\ \cline{2-8} 
\multicolumn{1}{|l|}{}                                  & \multicolumn{1}{c|}{ENZYMES}          & \multicolumn{1}{c|}{\cite{borgwardt2005protein}}     & \multicolumn{1}{c|}{600}            & \multicolumn{1}{c|}{6}               & \multicolumn{1}{c|}{32.63}                & \multicolumn{1}{c|}{62.14}                & \multicolumn{1}{c|}{\cite{atamna2019spi,nikolentzos2017matching,yanardag2015deep}}                            \\ \cline{2-8} 
\multicolumn{1}{|l|}{}                                  & \multicolumn{1}{c|}{PROTEINS}         & \multicolumn{1}{c|}{\cite{borgwardt2005protein}}     & \multicolumn{1}{c|}{1113}           & \multicolumn{1}{c|}{2}               & \multicolumn{1}{c|}{39.06}                & \multicolumn{1}{c|}{72.82}                & \multicolumn{1}{c|}{\cite{du2019graph,narayanan2017graph2vec,nikolentzos2017matching,yanardag2015deep}}                            \\ \hline
\multicolumn{1}{|c|}{\multirow{5}{*}{Chemoinformatics}} & \multicolumn{1}{c|}{AIDS}             & \multicolumn{1}{c|}{\cite{riesen2008iam}}            & \multicolumn{1}{c|}{2000}           & \multicolumn{1}{c|}{2}               & \multicolumn{1}{c|}{15.69}                & \multicolumn{1}{c|}{16.20}                & \multicolumn{1}{c|}{\cite{bai2018convolutional,bai2019simgnn,anonymous2019_matching}}                            \\ \cline{2-8} 
\multicolumn{1}{|c|}{}                                  & \multicolumn{1}{c|}{MUTAG}            & \multicolumn{1}{c|}{\cite{debnath1991structure}}     & \multicolumn{1}{c|}{188}            & \multicolumn{1}{c|}{2}               & \multicolumn{1}{c|}{17.93}                & \multicolumn{1}{c|}{19.79}                & \multicolumn{1}{c|}{\cite{al2019ddgk,du2019graph,atamna2019spi,narayanan2017graph2vec,nikolentzos2017matching,anonymous2019_inductive,yanardag2015deep}}                            \\ \cline{2-8} 
\multicolumn{1}{|c|}{}                                  & \multicolumn{1}{c|}{NCI1}             & \multicolumn{1}{c|}{\cite{wale2008comparison}}       & \multicolumn{1}{c|}{4110}           & \multicolumn{1}{c|}{2}               & \multicolumn{1}{c|}{29.87}                & \multicolumn{1}{c|}{32.30}                & \multicolumn{1}{c|}{\cite{al2019ddgk,atamna2019spi,du2019graph,nikolentzos2017matching,anonymous2019_inductive,yanardag2015deep}}                            \\ \cline{2-8} 
\multicolumn{1}{|c|}{}                                  & \multicolumn{1}{c|}{NCI109}           & \multicolumn{1}{c|}{\cite{wale2008comparison}}       & \multicolumn{1}{c|}{4127}           & \multicolumn{1}{c|}{2}               & \multicolumn{1}{c|}{29.68}                & \multicolumn{1}{c|}{32.13}                & \multicolumn{1}{c|}{\cite{narayanan2017graph2vec,nikolentzos2017matching,yanardag2015deep}}                            \\ \cline{2-8} 
\multicolumn{1}{|c|}{}                                  & \multicolumn{1}{c|}{PTC\_MR}          & \multicolumn{1}{c|}{\cite{helma2001predictive}}      & \multicolumn{1}{c|}{344}            & \multicolumn{1}{c|}{2}               & \multicolumn{1}{c|}{14.29}                & \multicolumn{1}{c|}{14.69}                & \multicolumn{1}{c|}{\cite{narayanan2017graph2vec,nikolentzos2017matching,yanardag2015deep}}                            \\ \hline
\multicolumn{1}{|c|}{\multirow{3}{*}{Brain Networks}}   & \multicolumn{1}{c|}{ABIDE}            & \multicolumn{1}{c|}{\cite{di2014autism}}                                                 & \multicolumn{1}{c|}{871}            & \multicolumn{1}{c|}{2}               & \multicolumn{1}{c|}{110}                  & \multicolumn{1}{c|}{-}                    & \multicolumn{1}{c|}{\cite{ktena2018metric,ma2019similarity}}                            \\ \cline{2-8} 
\multicolumn{1}{|l|}{}                                  & \multicolumn{1}{c|}{UK Biobank}       & \multicolumn{1}{c|}{\cite{biobank2014uk}}                                                 & \multicolumn{1}{c|}{2500}           & \multicolumn{1}{c|}{2}               & \multicolumn{1}{c|}{55}                   & \multicolumn{1}{c|}{-}                    & \multicolumn{1}{c|}{\cite{ktena2018metric}}                            \\ \cline{2-8} 
\multicolumn{1}{|l|}{}                                  & \multicolumn{1}{c|}{HCP}              & \multicolumn{1}{c|}{\cite{van2012human}}                                                 & \multicolumn{1}{c|}{1200}            & \multicolumn{1}{c|}{2}               & \multicolumn{1}{c|}{360}                  & \multicolumn{1}{c|}{-}                    & \multicolumn{1}{c|}{\cite{ma2019similarity}}                            \\ \hline
\multicolumn{1}{|c|}{Image Graphs}                      & \multicolumn{1}{c|}{COIL-DEL}         & \multicolumn{1}{c|}{\cite{riesen2008iam}}                                                 & \multicolumn{1}{c|}{3900}           & \multicolumn{1}{c|}{100}             & \multicolumn{1}{c|}{21.54}                & \multicolumn{1}{c|}{54.24}                & \multicolumn{1}{c|}{\cite{li2019graph}}                            \\ \hline
                                                        & \multicolumn{1}{l}{}                  &                                                                       & \multicolumn{1}{l}{}                & \multicolumn{1}{l}{}                 & \multicolumn{1}{l}{}                      & \multicolumn{1}{l}{}                      &                                                 
\end{tabular}
}
\end{center}
\end{table*}

\gm{In this section, we summarize the characteristics of the datasets that are frequently used in deep graph similarity learning methods and the experimental evaluation adopted by these methods. }

\subsection{Datasets}
\gm{Graph data from various domains have been used to evaluate graph similarity learning methods~\cite{rossi2015network}, for example, protein-protein graphs from bioinformatics, chemical compound graphs from chemoinformatics, and brain networks from neuroscience, etc. We summarize the benchmark datasets that are frequently used in deep graph similarity learning methods in Table~\ref{tab:datasets}. }

\gm{In addition to these datasets, synthetic graph datasets or other domain-specific datasets are also widely used in some graph similarity learning works. For example, in \cite{li2019graph} and \cite{anonymous2019_matching}, control flow graphs of binary functions are generated and used to evaluate graph matching networks for binary code similarity search. In \cite{ijcai2019-522}, attacks are conducted on testing machines to generate malware data, which are then merged with normal data to evaluate the Siamese GNN model for malware detection. In \cite{jiang2019glmnet}, images are collected from multiple categories and keypoints are annotated in the images to evaluate the proposed model for graph matching.}

\subsection{Evaluation}
\gm{During evaluation, most GSL methods take pairs or triplets of graphs as input during training with various objective functions used for various graph similarity tasks. The existing evaluation tasks mainly include pair classification ~\cite{xu2017neural,ktena2018metric,ma2019similarity,li2019graph,anonymous2019_matching}, graph classification~\cite{tixier2019graph,nikolentzos2017matching,narayanan2017graph2vec,atamna2019spi,wu2018dgcnn,anonymous2019_inductive,liu2019n,yanardag2015deep,al2019ddgk,du2019graph}, graph clustering~\cite{anonymous2019_inductive}, graph distance prediction~\cite{bai2018convolutional,bai2019simgnn,anonymous2019_matching}, and graph similarity search~\cite{ijcai2019-522}. Classification AUC (i.e., Area Under the ROC Curve) or accuracy are used as the most popular metric for the evaluation of graph-pair classification or graph classification task \cite{ma2019similarity,li2019graph}. Mean squared error (MSE) is used as evaluation metric for the regression task in graph distance prediction~\cite{bai2018convolutional,bai2019simgnn}.}

\gm{According to the evaluation results reported in the above works, the deep graph similarity learning methods tend to outperform the traditional methods. For example, \cite{al2019ddgk} shows that the deep divergence graph kernel approach achieves higher classification accuracy scores compared to traditional graph kernels such as the shortest-path kernel \cite{borgwardt2005shortest} and Weisfeiler-Lehman kernel \cite{kriege2016valid} in most cases for the graph classification task. Meanwhile, among the deep methods, methods that allow for cross-graph feature interaction tend to achieve a better performance compared to the factorized methods that relies only on single graph features. For instance, the experimental evaluations in \cite{li2019graph} and \cite{anonymous2019_matching} have demonstrated that the GNN-based graph matching networks have a superior performance than the Siamese GNNs in pair classification and graph edit distance prediction tasks. }

\gm{The efficiency of different methods are also analyzed and evaluated in some of these works.  In \cite{bai2019simgnn}, some evaluations have been done for comparing the efficiency of the GNN based graph similarity learning approach SimGNN with traditional GED approximation methods including A*-Beamsearch \cite{neuhaus2006fast}, Hungarian \cite{riesen2009approximate} and VJ \cite{fankhauser2011speeding}, where the core operation for GED approximation may take polynomial or sub-exponential to the number of nodes in the graphs. For the GNN based model like SimGNN, to compute similarity score for pairs of graphs, the time complexity mainly involves two parts: (1) the node-level and graph-level embedding computation stages, where the time complexity is $O(|E|)$, and $|E|$ is the number of edges of the graph \cite{kipf2016semi}; and (2) the similarity score computation stage, where the time complexity is $O(D^2K)$ ($D$ is the dimension of the graph-level embedding, and $K$ is the feature map dimension used in the graph-graph interaction stage) for the strategy of using graph-level embedding interaction, and the time complexity is $O(DN^2)$ ($N$ is the number of nodes in the larger graph). The experimental evaluations in \cite{bai2019simgnn} show that the GNN based models consistently achieve the best results in efficiency and effectiveness for the pairwise GED computation \cite{bai2019simgnn} on multiple graph datasets, demonstrating the benefit of using these deep models for the similarity learning tasks.}  

\section{Applications}\label{sec:apps}
Graph similarity learning is a fundamental problem in domains where data are represented as graph structures, and it has various applications in the real world. 
\subsection{Computational Chemistry and Biology}
An important application of graph similarity learning in the chemistry and biology domain is to learn the chemical similarity, which aims to learn the similarity of chemical elements, molecules or chemical compounds with respect to their effect on reaction partners in inorganic or biological settings \cite{brown2009chemoinformatics}. An example is the compounds query for in-silico drug screening, where searching for similar compounds in a database is the key process. 

In the literature of graph similarity learning, quite a number of models have been proposed and applied to similarity learning for chemical compounds or molecules. Among these work, the traditional models mainly employ sub-graph based search strategies or graph kernels to solve the problem \cite{zheng2013graph,zeng2009comparing,swamidass2005kernels,mahe2009graph}. However, these methods tend to have high computational complexity and strongly rely on the sub-graph or kernels defined, making it difficult to use them in real applications. Recently, a deep graph similarity learning model SimGNN is proposed in \cite{bai2019simgnn} which also aims to learn similarity for chemical compounds as one of the tasks. Instead of using sub-graphs or other explicit features, the model adopts GCNs to learn node-level embeddings, which are fed into an attention module after multiple layers of GCNs to generate the graph-level embeddings. Then a neural tensor network (NTN) \cite{socher2013reasoning} is used to model the relation between two graph-level embeddings, and the output of the NTN is used together with the pairwise node embedding comparison output in the fully connected layers for predicting the graph edit distance between the two graphs. This work has shown that the proposed deep learning model outperforms the traditional methods for graph edit distance computation in prediction accuracy and with much less running time, which indicates the promising application of the deep graph similarity learning models in the chemo-informatics and bio-informatics. 

%tries to learn graph embedding from the graphs and leverage that in the similarity learning process to make it an end-to-end framework for similarity score prediction.  

\subsection{Neuroscience}
\label{sec:apps_neuro}
Many neuroscience studies have shown that structural and functional connectivity of the human brain reflects the brain activity patterns that could be indicators of the brain health status or cognitive ability level \cite{badhwar2017resting,ma2017multi,ma2017multib}. For example, the functional brain connectivity networks derived from fMRI neuroimaging data can reflect the functional activity across different brain regions, and people with brain disorder like Alzheimer's disease or bipolar disorder tend to have functional activity patterns that differ from those of healthy people \cite{badhwar2017resting,syan2018resting,ma2016multi}. To investigate the difference in brain connectivity patterns for these neuroscience problems, researchers have started to study the similarity of brain networks among multiple subjects with graph similarity learning methods \cite{lee2020deep,ktena2018metric,ma2019similarity}.

The organization of functional brain networks is complicated and usually constrained by various factors, such as the underlying brain anatomical network, which plays an important role in shaping the activity across the brain. These constraints make it a challenging task to characterize the structure and organization of brain networks while performing similarity learning on them. Recent work in \cite{ktena2018metric}, \cite{ma2019similarity} and \cite{liu2019community} have shown that the deep graph models based on graph convolutional networks have a superior ability to capture brain connectivity features for the similarity analysis compared to the traditional graph embedding based approaches. In particular, \cite{ma2019similarity} proposes a higher-order Siamese GCN framework that leverages higher-order connectivity structure of functional brain networks for the similarity learning of brain networks. 

\gm{In view of the work introduced above and the trending research problems in the field of neuroscience, we believe that deep graph similarity learning will benefit the clinical investigation of many brain diseases and other neuroscience applications. Promising research directions include, but are not limited to, deep similarity learning on resting-state or task-related fMRI brain networks for multi-subject analysis with respect to brain health status or cognitive abilities, deep similarity learning on the temporal or multi-task fMRI brain networks of individual subjects for within-subject contrastive analysis over time or across tasks for neurological disorder detection. Some example fMRI brain network datasets that can be used for such analysis have been introduced in Table 3.} 

\subsection{Computer Security}
In the field of computer security, graph similarity has also been studied for various application scenarios, such as the hardware security problem \cite{marc2019}, the malware indexing problem based on function-call graphs \cite{hu2009large}, and the binary function similarity search for identifying vulnerable functions \cite{li2019graph}. 

In \cite{marc2019}, a graph similarity heuristic is proposed based on spectral analysis of adjacency matrices for the hardware security problem, where evaluations are done for three tasks, including gate-level netlist reverse engineering, Trojan detection, and obfuscation assessment. The proposed method outperforms the graph edit distance approximation algorithm proposed in \cite{hu2009large} and the neighbor matching approach \cite{vujovsevic2013software}, which matches neighboring vertices based on graph topology. \cite{li2019graph} is the work that introduced GNN-based deep graph similarity learning models to the security field to solve the binary function similarity search problem. Compared to previous models, the proposed deep model computes similarity scores jointly on pairs of graphs rather than first independently mapping each graph to a vector, and the node representation update process uses an attention-based module which considers both within-graph and cross-graph information. Empirical evaluations demonstrate the superior performance of the proposed deep graph matching networks compared to the Google's open source function similarity search tool \cite{functionsim}, the basic GNN models, and the Siamese GNNs. 
\subsection{Computer Vision}
\label{sec:apps_cv}
Graph similarity learning has also been explored for applications in computer vision. In \cite{wu2014human}, context-dependent graph kernels are proposed to measure the similarity between graphs for human action recognition in video sequences. Two directed and attributed graphs are constructed to describe the local features with intra-frame relationships and inter-frame relationships, respectively. The graphs are decomposed into a number of primary walk groups with different walk lengths, and a generalized multiple kernel learning algorithm is applied to combine all the context-dependent graph kernels, which further facilitates human action classification. In \cite{guo2018neural}, a deep model called Neural Graph Matching Network is first introduced for the 3D action recognition problem in the few-shot learning setting. Interaction graphs are constructed from the 3D scenes, where the nodes represent physical entities in the scene and edges represent interactions between the entities. The proposed NGM Networks jointly learn a graph generator and a graph matching metric function in an end-to-end fashion to directly optimize the few-shot learning objective. It has been shown to significantly improve the few-shot 3D action recognition over the holistic baselines. 

\gm{Another emerging application of graph similarity learning in computer vision is the image matching problem, where the goal is to find consistent correspondences between the sets of features in two images. As introduced at the end of Section 3.2, recently some deep graph matching networks have been developed for the image matching task \cite{jiang2019glmnet,wang2019learning}, where images are first converted to graphs and the image matching problem is then solved as a graph matching problem. In the graph converted from an image, the nodes represent the unary descriptors of annotated feature points in images, and edges encode the pairwise relationships among different feature points in that image. Based on the new graph representation, the feature matching can be reformulated as graph matching problem. However, it is worth noting that, this graph matching is actually the graph node matching, as the goal is to match the nodes between graphs instead of two entire graphs. Therefore, the graph based image matching problem is a special case or a sub-problem of the general graph matching problem. }

\gm{The two application problems discussed above are both promising directions of applying deep graph similarity learning models for the practical learning tasks in computer vision. A key advice we provide on applying graph similarity learning methods for these image applications is to first find an appropriate mapping for converting the images to graphs, so that the learning tasks on images can be formulated as the graph similarity learning based tasks.}

\section{Challenges}\label{sec:challenges}
\subsection{Various Graph Types}
In most of the work discussed above, the graphs involved consist of unlabeled nodes/edges and undirected edges. However, there are many variants of graphs in real world applications. How to build deep graph similarity learning models for these various graph types is a challenging problem. 
\noindent \paragraph{Directed Graphs.}
In some application scenarios, the graphs are directed, which means all the edges in the graph are directed from one vertex to another. For instance, in a knowledge graph, edges go from one entity to another, where the relationship is directed. In such cases, we should treat the information propagation process differently according to the direction of the edge. Recently some GCN based graph models have suggested some strategies for dealing with such directed graphs. In \cite{kampffmeyer2019rethinking}, a dense graph propagation strategy is proposed for the propagation on knowledge graphs, where two kinds of weight matrices are introduced for the propagation based on a node's relationship to its ancestors and descendants respectively. However, to the best of our knowledge, no work has been done on deep similarity learning specifically for directed graphs, which arises as a challenging problem for this community.
\noindent \paragraph{Labeled Graphs.}
Labeled graphs are graphs where vertices or edges have labels. For example, in chemical compound graphs where vertices denote the atoms and the edges represent the chemical bonds between the atoms, each node and edge have labels representing the atom type and bond type, respectively. These labels are important for characterizing the node-node relationship in the graphs, therefore it is important to leverage these label information for the similarity learning. In \cite{bai2019simgnn,ahmed2018learning}, the node label information are used as the initial node representations encoded by a one-hot vector and used in the node embedding stage. In this case, the nodes with same type share the same one-hot encoding vector. This should guarantee that even if the node ids are permuted, the aggregation results would be the same. However, the label information is only used for the node embedding process within each graph, and the comparison of the node or edge labels across graphs is not considered during the similarity learning stage. \gm{In \cite{al2019ddgk}, both node labels and edge labels in the chemo- and bio-informatic graphs have been used as attributes for learning better alignment across graphs, which has been shown to lead to a better performance. Therefore, how to leverage the node / edge attributes of the labeled graphs into the similarity learning process is a critical problem.}  

\noindent \paragraph{Dynamic and Streaming Graphs.}
\label{sec:chall_dynamic}
Another type of graphs is the dynamic graph, which has a static graph structure and dynamic input signals/features. For example, the 3D human action or motion data can be represented as graphs where the entities are represented as nodes and the actions as edges connecting the entities. Then similarity learning on these graphs is an important problem for action and motion recognition. Moreover, another type of graph is the streaming graph, where both the structure and/or features are continuously changing~\cite{ahmed2019temporal,ahmed2019network}. For example, online social networks~\cite{ahmed2017sampling,ahmed2014graph,ahmed2014network}. The similarity learning would be important for change/anomaly detection, link prediction, relationship strength prediction, etc. Although some work has proposed variants of GNN models for spatio-temporal graphs \cite{yu2017spatio,manessi2020dynamic}, and other learning methods for dynamic graphs~\cite{nguyen2018continuous,nguyen2018dynamic,tong2008colibri,li2017attributed}, the similarity learning problem on dynamic and streaming graphs has not been well studied. \gm{For example, in the multi-subject analysis of task-related fMRI brain networks as mentioned in Section~\ref{sec:apps_neuro}, for each subject, a set of brain connectivity networks can be collected for a give time period, which forms a spatio-temporal graph. It would be interesting to conduct similarity learning on the spatio-temporal graphs of different subjects to analyze their similarity in cognitive abilities, which is an important problem in the neuroscience field. However, to the best of our knowledge, none of the existing similarity learning methods is able to deal with such spatio-temporal graphs.} The main challenge in such problems is how to leverage the temporal updates of the node-level representations and the interactions between the nodes on these graphs while modeling their similarity. 

\subsection{Interpretability}
The deep graph models, such as GNNs, combine node feature information with graph structure by recursively passing neural messages along edges of the graph, which is a complex process and makes it challenging to explain the learning results from these models. Recently, some work has started to explore the interpretability of GNNs \cite{ying2019gnn,baldassarre2019explainability}. In \cite{ying2019gnn}, a GNNEXPLAINER is proposed for providing interpretable explanations for predictions of GNN-based models. It first identifies a subgraph structure and a subset of node features that are crucial in a prediction. Then it formulates an optimization task that maximizes the mutual information between a GNN's prediction and the distribution of possible subgraph structures. \cite{baldassarre2019explainability} explores the explainability of GNNs using gradient-based and decomposition-based methods, respectively, on a toy dataset and a chemistry task. Although these works have provided some insights into the interpretability of GNNs, they are mainly for node classification or link prediction tasks on a graph. To the best of our knowledge, the explainability of GNN-based graph similarity models remains unexplored.

\subsection{Few-shot Learning}
The task of few-shot learning is to learn classifiers for new classes with only a few training examples per class. A big branch of work in this area is based on metric learning \cite{wang2019few}. However, most of the existing work proposes few-shot learning problems on images, such as image recognition \cite{koch2015siamese} and image retrieval\cite{triantafillou2017few}. Little work has been done on metric learning for few-shot learning on graphs, which is an important problem for areas in which data are represented as graphs and data gathering is difficult, for example, brain connectivity network analysis in neuroscience. Since graph data usually has complex structure, how to learn a metric so that it can facilitate generalizing from a few graph examples is a big challenge. Some recent work ~\cite{guo2018neural} has begun to explore the few-shot 3D action recognition problem with graph-based similarity learning strategies, where a neural graph matching network is proposed to jointly learn a graph generator and a graph matching metric function to optimize the few-shot learning objective of 3D action recognition. However, since the objective is defined specifically based on the 3D action recognition task, the model can not be directly used for other domains. The remaining problem is to design general deep graph similarity learning models for the few-shot learning task for a multitude of applications. 

\section{Conclusion} \label{sec:conc}
Recently, there has been an increasing interest in deep neural network models for learning graph similarity. In this survey paper, we provided a comprehensive review of the existing work on deep graph similarity learning, and categorized the literature into three main categories: (1) graph embedding based graph similarity learning models, (2) GNN-based models, and (3) Deep graph kernels. We discussed and summarized the various properties and applications of the existing literature. Finally, we pointed out the key challenges and future research directions for the deep graph similarity learning problem. 

\section{Acknowledgement}
Philip S. Yu is supported by NSF under grants III-1526499, III-1763325, III-1909323, and SaTC-1930941.

\bibliographystyle{plain}

% \begin{abstract}
% Insert your abstract here. Include keywords, PACS and mathematical
% subject classification numbers as needed.
% \keywords{First keyword \and Second keyword \and More}
% % \PACS{PACS code1 \and PACS code2 \and more}
% % \subclass{MSC code1 \and MSC code2 \and more}
% \end{abstract}

%\begin{acknowledgements}
%\end{acknowledgements}



\end{document}